\documentclass{article}

\usepackage[preprint]{corl_2023} 

\title{A Population-Level Analysis of Neural Dynamics in Robust Legged Robots}

\author{
  Eugene R. Rush\thanks{Correspondence to \href{mailto:eugene.rush@colorado.edu}{eugene.rush@colorado.edu}} \hspace{10pt} Christoffer Heckman \hspace{10pt} Kaushik Jayaram \hspace{10pt} J. Sean Humbert \\
  \\
  University of Colorado Boulder
}

\usepackage{graphicx} 
\usepackage{subfig} 
\usepackage{amsmath} 
\usepackage[export]{adjustbox} 

\begin{document}
\maketitle



\begin{abstract}

Recurrent neural network-based reinforcement learning systems are capable of complex motor control tasks such as locomotion and manipulation, however, much of their underlying mechanisms still remain difficult to interpret. Our aim is to leverage computational neuroscience methodologies to understanding the population-level activity of robust robot locomotion controllers. Our investigation begins by analyzing topological structure, discovering that fragile controllers have a higher number of fixed points with unstable directions, resulting in poorer balance when instructed to stand in place. Next, we analyze the forced response of the system by applying targeted neural perturbations along directions of dominant population-level activity. We find evidence that recurrent state dynamics are structured and low-dimensional during walking, which aligns with primate studies. Additionally, when recurrent states are perturbed to zero, fragile agents continue to walk, which is indicative of a stronger reliance on sensory input and weaker recurrence.

\end{abstract}

\keywords{Neuroscience, Reinforcement Learning, Motor Control, Locomotion} 


\section{Introduction}

In recent years, deep reinforcement learning (RL) systems have demonstrated intelligent behavior in real-world settings, especially in the areas of locomotion \cite{rudin_learning_2022,lee_learning_2020,rudin_advanced_2022,miki_learning_2022,siekmann_learning_2020,siekmann_sim--real_2021,vollenweider_advanced_2022,feng_genloco_2022} and manipulation \cite{gu_deep_2016,zeng_learning_2018,kalashnikov_qt-opt_2018,openai_solving_2019,fu_deep_2022}. These accomplishments not only advance the field, but also represent significant steps towards generating robot behavior as complex as animals. In fact, agility benchmarks have been established between quadrupedal robots and their furry counterparts \cite{caluwaerts_barkour_2023}. 

Some of these RL-based controllers employ artificial recurrent neural networks (RNNs), rather than frame stacking in order to capture sensory information that is not contained within a single time step, as well as unobserved environmental information. However, few papers have studied the dynamics within these RNN-based controllers during embodied closed-loop tasks. Within the past decade, a `computation through dynamics' (CTD) paradigm \cite{sussillo_opening_2013,saxena_towards_2019,vyas_computation_2020} has emerged in the computational neuroscience community, where inherent dynamical system structure is found in high-dimensional neural populations. These dynamics have been characterized through topological analysis of fixed points \cite{golub_fixedpointfinder_2018,maheswaranathan_universality_2019}, as well as trajectory-based geometric analysis \cite{russo_motor_2018,russo_neural_2020,saxena_motor_2022}.

Recently, \cite{oshea_direct_2022} utilized CTD methodologies to better understand the underlying computational principles of robust motor control in primates. Motor neuroscientists have three competing hypotheses: that networks have (a) high-dimensional reservoir dynamics, (b) low-dimensional subspace structured dynamics, or (c) path-following dynamics. After training primates on a reaching task, researchers in this study applied targeted neural perturbations, and found evidence for structured dynamics operating in low-dimensional subspaces. Despite the success of this work, there has yet to be any equivalent studies that interrogate the relationship between RNN dynamics and robust performance in RL-based motor control systems.

This work aims to apply these powerful CTD methodologies to robotics and reinforcement learning applications, in order to develop a deeper understanding of movement control in these learned RNN-based control systems. Specifically, we study the neural activity of RNNs within RL-based controllers, and the resulting agent behavior during motor control tasks. This work proposes the following contributions, which are unique in the context of robust motor control of a recurrent RL-based quadrupedal robot:
\begin{enumerate}
    \item Fragile RNN controllers have more saddle points, with at least one unstable direction
    \item More complex topology of fragile controllers leads to more stepping in place, whereas robust controllers balance statically
    \item Targeted neural perturbations reveal structured low-dimensional recurrent dynamics seen in primates
\end{enumerate}



\section{Related Work}
\label{sec:related_work}

\textbf{Solving cognitive tasks using RNNs.} Many studies in recent years have trained RNNs to solve cognitive tasks, ranging from text classification \cite{aitken_geometry_2022} and sentiment analysis \cite{maheswaranathan_how_2020,maheswaranathan_reverse_2019}, to transitive inference \cite{kay_neural_2022}, pose estimation \cite{cueva_emergence_2018,cueva_emergence_2020,cueva_recurrent_2021}, memory \cite{cueva_low-dimensional_2020}, and other cognitive tasks \cite{yang_task_2019}. With direct access to the full parameterization of these models, computational neuroscientists have made efforts to reverse-engineer these systems. However, these cognitive studies are inherently open-loop estimators, and do not translate to closed-loop control tasks, such as motor control.

\textbf{Interpreting embodied RL systems.} In contrast, RL is a natural solution for connecting sensorimotor processing with goal-directed, embodied behavior. Despite this, few studies have brought a CTD perspective to closed-loop control tasks. One exception is \cite{merel_deep_2020}, in which researchers simulate a virtual rodent model and study neural activity across various high-level behaviors. Another is an in-silico study \cite{singh_emergent_2021} that examines the population-level dynamics of a virtual insect localizing and navigating to the source of an an odor plume. The analyses in these studies reveal coordinated neural activity patterns, however neither make direct connections to motor neuroscience hypotheses: the former \cite{merel_deep_2020} focusing chiefly on features of multi-task neural behavior, and the latter \cite{singh_emergent_2021} focusing on how neural activity relates to spatial localization and navigation. Lastly, a recent study inspired from motor neuroscience has found smooth, untangled neural trajectories within RNN-based locomotion controllers that are similar to those found in primates \cite{rush_data-fitting_2023}. However, this study does not explore how this neural phenomenon relates to robust behavioral performance.


\textbf{Motor control in robotics and RL.} In recent years the robotics community has demonstrated increasingly complex task-oriented behavior, such as legged locomotion \cite{rudin_learning_2022,lee_learning_2020,miki_learning_2022,feng_genloco_2023} and dexterous manipulation \cite{openai_learning_2019,handa_dextreme_2022}, but has a sparse literature on relating neural activity to embodied behavior. This is largly because many high-performance RL systems opt out of utilizing RNNs, which are often less straightforward to train. These feedforward systems typically satisfy the Markov assumption -- that observation states completely describe the system -- through frame stacking, where computer memory is used to store prior observations with current ones. For systems with periodic reward functions, an alternate solution is to provide a clock signal from the computer as a temporal reference \cite{siekmann_sim--real_2021}.

There is however, a subset of RL-based robotic motor controllers that has found major success in employing RNNs, \cite{siekmann_learning_2020,rudin_learning_2022,openai_learning_2019,openai_solving_2019,handa_dextreme_2022}. While there have been studies that explore the neural activity and behavior of in-silico RNN-based RL models  \cite{siekmann_learning_2020,merel_deep_2020,singh_emergent_2021}, none have focused on better understanding the robustness of motor control. Overall, it is clear that advancements in deep RL have led to functional breakthroughs in robotics, but remains fertile ground for explainability, especially through the lens of computational neuroscience and the CTD paradigm.

\section{Method}
\label{sec:method}

To elucidate salient properties of robust locomotion, we first train quadrupedal robotic agents to walk in a virtual environment, which is discussed in Section \ref{sec:experimental_setup}. Given an RNN-based controller and its embodied motor control behavior, how can we elucidate neural properties that relate to the agent's ability to reject disturbances?

\textbf{Unforced Characteristics.} Our first aim is to understand the unforced dynamics in the system. 
We begin by first finding and characterizing the fixed points of the RNN. To do this, we search the recurrent states in directions of slowest dynamics \cite{sussillo_opening_2013,golub_fixedpointfinder_2018}, utilizing stochastic gradient descent \cite{kingma_adam_2017}. To find these fixed points, random initial recurrent states are sampled, ran through the optimization, and after all points converge, clustered to unique states. The resulting fixed points are classified, by linearizing the dynamics around the fixed point, computing the recurrent state jacobians, and computing the eigenvalues of each fixed point.
 
\textbf{Neural Perturbations.} Topological analysis provides us an understanding of how recurrent states decay if sensory inputs are suddenly set to zero. However, in order to deepen our understanding of robust behavior, we need to examine the forced response, i.e. where the agent experiences sensory feedback. We apply neural perturbations during walking, with the aim of deepening our understanding of disturbance rejection properties of our locomotion controller during nominal operation.

We draw inspiration for experiment design from primate studies, which found low-dimensional structure in their population dynamics \cite{oshea_direct_2022}, and to date has not been applied to RL-based agents. The only similar RL study \cite{merel_deep_2020} perturbed RNN hidden states by inactivating neuronal states or replacing with neural states from other behavioral policies, but did not apply targeted perturbations to better understand low-dimensional structure.





\section{Experimental Setup}
\label{sec:experimental_setup}

\textbf{Agent and Environment.} For this work, we incorporate NVIDIA Isaac Gym \cite{makoviychuk_isaac_2021}, an end-to-end GPU-accelerated physics simulation environment, and a virtual model of the qudrupedal Anymal robot \cite{hutter_anymal_2016,rudin_learning_2022} from IsaacGymEnvs\footnote{\href{https://github.com/NVIDIA-Omniverse/IsaacGymEnvs}{https://github.com/NVIDIA-Omniverse/IsaacGymEnvs}}. The action space is continuous and consists of 12 motor torque commands, three for each leg. The observation space consists of 36 signals: three translational body velocities $(u, v, w)$, three rotational body velocities $(p, q, r)$, three body orientation angles $(\theta, \phi, \psi)$, three planar body velocity commands $(u^{*}, v^{*}, r^{*})$, 12-dimensional joint angle positions $\boldsymbol{\theta_\text{joint}}$, 12-dimensional joint angular velocity $\boldsymbol{\omega_\text{joint}}$. The agent utilizes 140 depth measurements during training as well, however they are ignored in this work, since we roll out policies on flat ground.

\textbf{Model Training.} The reward function consists of a weighted sum of linear body velocity error $(u^{*}-u)^{2} + (v^{*}-v)^{2}$ and angular body velocity error $p^2 + q^2 + (r^{*}-r)^2$, along with a suite of knee collision, joint acceleration, change in torque, and foot airtime penalties. During training, agents are randomly assigned linear body velocity commands $(u^{*},v^{*})$, in addition to an angular velocity command $r^{*}$ that is modulated to regulate to a random goal heading. Additionally, agents are trained using the AnymalTerrain curriculum learning, during which they gradually face more difficult terrain as training progresses. Agents are also exposed to sensory noise and slight randomimzations to gravity and friction during training.

\begin{figure}[h]
    \centering
    \includegraphics[width=0.65\textwidth]{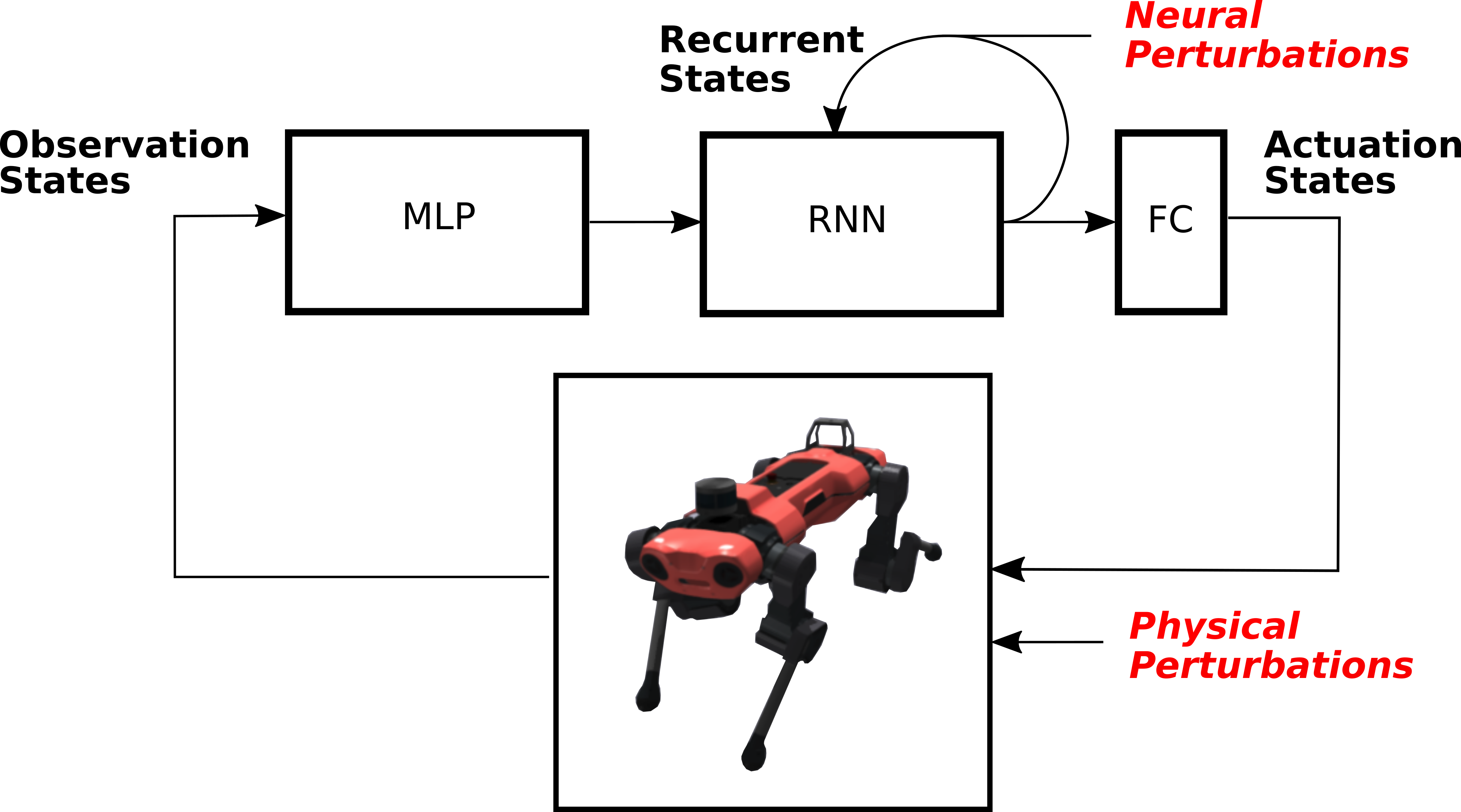}
    \caption{System diagram of RNN-based reinforcement learning locomotion controller. The observation states consists of body velocity commands and sensory feedback signals, which pass through a MLP and an RNN. The recurrent state feeds back on itself in the RNN, and part of it feeds into a fully-connected (FC) layer, and outputs the actuation state, which consists of 12 motor torque commands. As the output of the controller, these commands are sent to Anymal, which is modeled in the NVIDIA Isaac Gym physics simulator. The resulting observation states consist of sensory feedback signals such as body velocities, orientations, joint positions and velocities, as well as body velocity command signals. These are then updated and are fed back into the RL controller for the subsequent time step.}
    \label{fig:agent_environment}
\end{figure}

\textbf{Model Architecture.} The agent is trained using rl\_games\footnote{\href{https://github.com/Denys88/rl_games}{https://github.com/Denys88/rl\_games}}, a high performance RL libary that implements a variant of Proximal Policy Optimization (PPO) \cite{schulman_proximal_2017} that utilizes asymmetric inputs to actor and critic networks \cite{pinto_asymmetric_2017}. We utilize an implementation that integrates Long-Short Term Memory (LSTM) networks into both the actor and critic networks. Each actor and critic network passes the observation vector $[128 \times 1]$ through a multi-layer perceptron (MLP) with two dense layers of size 512 and 256, a single-layer, 128-cell LSTM network, and a fully-connected output layer that results in an action vector $[12 \times 1]$ as shown in Figure \ref{fig:agent_environment}. The neural activations of these cell states $[128 \times 1]$ and hidden states $[128 \times 1]$, are collectively referred to as the recurrent states $[256 \times 1]$. The action vector contains the motor torque commands for each of the 12 joints, and is referred to as the actuation state. The critic network is not illustrated, but has independently trained weights and identical structure, save the fully-connected layer which outputs a scalar value estimate, as opposed to as 12-dimensional actuation state.



\textbf{Training for Robustness.} The rl\_games implementation utilizes truncated backprogogation through time (BPTT) for training the RNNs. With a truncation length of four, this essentially unfolds the RNN into four layers, upon which gradients are propagated back through the network for each time step. Generally, BPTT truncation aids in training because it prevents long sequences from causing vanishing or exploding gradient issues, yet shorter truncations can limit the model's capacity for learning temporal patterns. 

\begin{figure}[h!]
    \centering
    \includegraphics[width=0.75\textwidth]{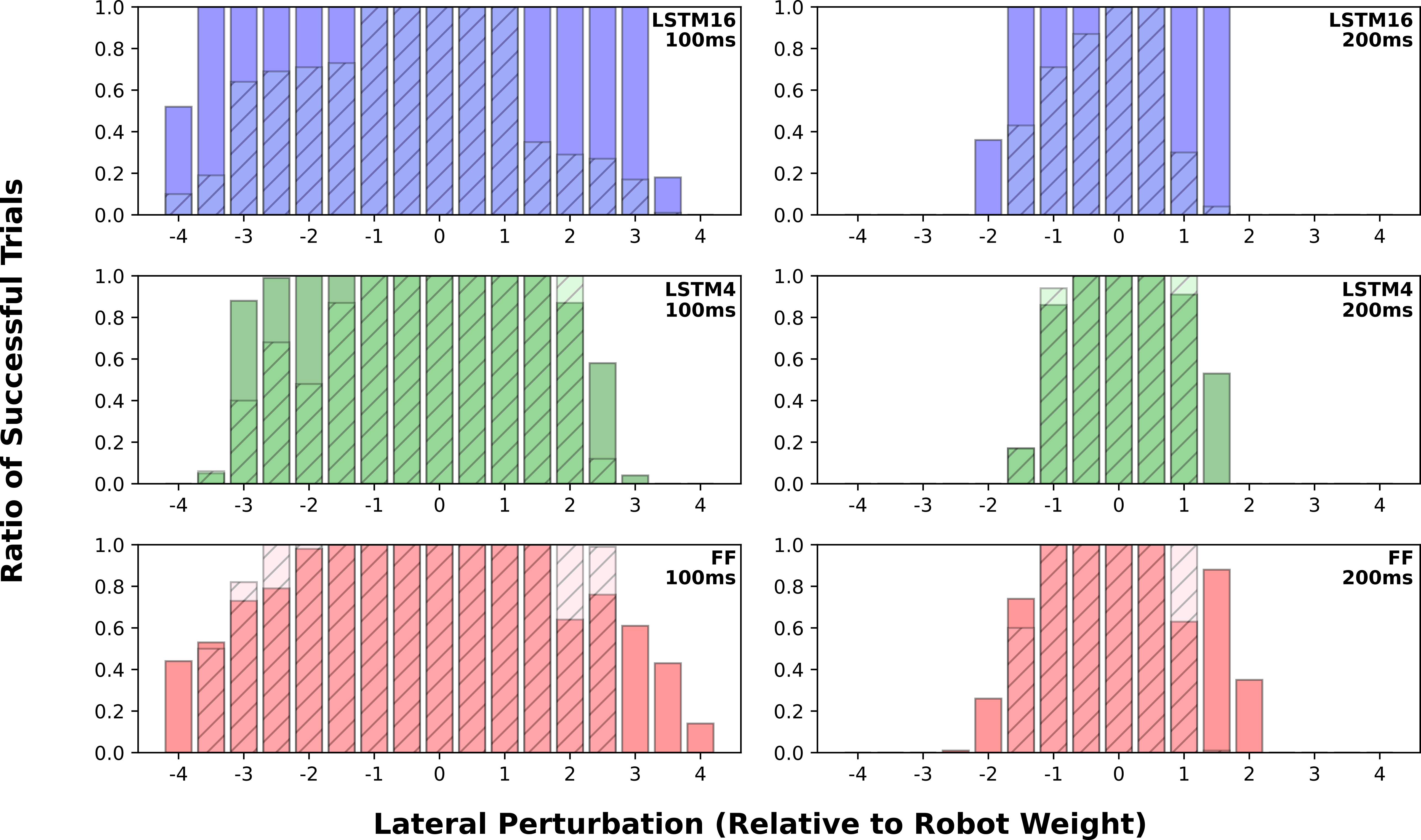}
    \caption{To quantify robustness to disturbances, each robot model (LSTM16, LSTM4, and FF) is evaluated by providing external perturbations of varying magnitude (-4x to 4x body weight) and duration (100ms and 200ms). The LSTM16 model is the most robust, reliably recovering from perturbations of 3x to 3.5x body weight. We evaluate the feedforward-only model purely for reference, and find that the LSTM4 model underperforms relative to the FF benchmark. Note that each column in this chart represents the percentage of agents (N=100) that recover from the surprise perturbation. Each of the three models have two sub-varieties, ones that receive perturbations during training (solid bars) and ones that do not (hatched bars). For each of these three models and two model sub-varieties, we simulate 100 agents over 17 different perturbation magnitudes and 2 different perturbation durations, resulting in a total of 20,400 individual trials.}
    \label{fig:robustness_statistics}
\end{figure}

We trained three models, one with a BPTT truncation length of 16 (LSTM16), one with a BPTT truncation length of 4 (LSTM4), and a feedforward-only model (FF) as a reference benchmark. We evaluate these three controller models by applying lateral perturbations of varying magnitude and duration during forward walking, and quantify the percentage of agents that successfully recover. From these three models, we find that the FF model outperforms the LSTM16 and LSTM4 models, as shown by the hatched-formatted data in Figure \ref{fig:robustness_statistics}.

A subsequent variety of these three models are trained, in which agents are exposed to random perturbations throughout the training procedure. Receiving these random perturbations during training modestly improves LSTM4 robustness, and significantly improves LSTM16 robustness, as shown by the solid-patterned data in Figure \ref{fig:robustness_statistics}. This is significant, because the random perturbations during training are different from the targeted lateral perturbations reflected in Figure \ref{fig:robustness_statistics}. During training, there is a 1\% chance of a perturbation being initiated each time step, and once that perturbation is initiated, there is a 2\% chance of termination each time step. During that  duration, the perturbation is held constant, where the x-, y-, and z-direction magnitudes are sampled from a uniform distribution between -0.23x to 0.23x body weight. Therefore, it is impressive that the LSTM4 model can reliably reject disturbances of -2.5x to 1.5x body weight, and the LSTM16 model can reliably reject disturbances of -3.5x to 3x body weight. Being able to adapt to order-of-magnitude greater perturbations is in alignment with other emergent meta-learning that LSTMs have been shown to exhibit \cite{openai_solving_2019}. We aim to elucidate the computational principles that drive this robust behavior in LSTM16, relative to LSTM4, and do so from analyzing its neural dynamics.

\textbf{Dimensionality Reduction.} After training and during data collection rollouts, the agent is commanded with a range of forward speed commands $u^{*}$ between 0.8 m/s and 2 m/s, while $v^{*}$ and $r^{*}$ remain at $0$ m/s. We perform principal component analysis (PCA) on this dataset, in order to determine the dominant neural populations and improve interpretability. For all perturbation studies presented in this work, we hold the forward speed command $u^{*}$ at 2 m/s, and then transform the data data based on speed-modulated PCA transformation. During data collection, noise parameters and perturbations are removed, unless otherwise stated.

\section{Results}

\subsection{Unforced Characteristics}

\begin{figure}[h]
    \centering
    \includegraphics[valign=c,width=0.4\textwidth]{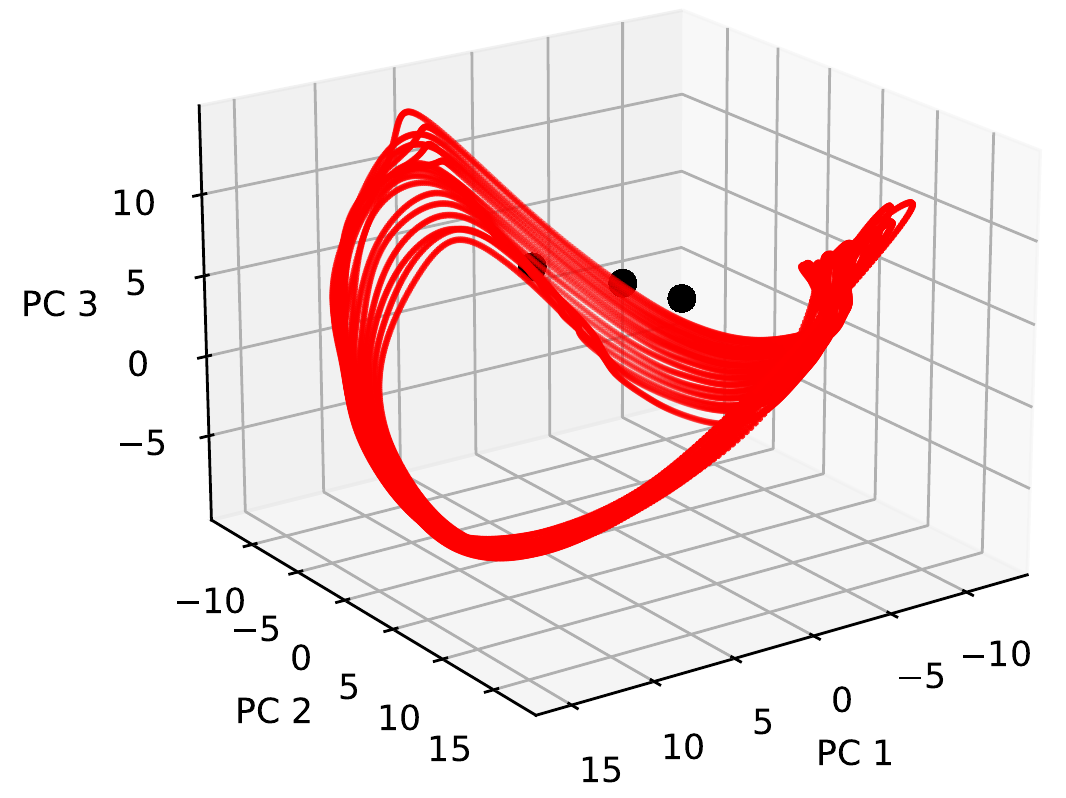}    
    \caption{LSTM16 has three fixed points, which are residing in the center of the cyclic recurrent state trajectories during forward walking (0.8 to 2 m/s).}
    \label{fig:3d_LSTM16}
\end{figure}

\begin{figure}[h]
    \centering
    \includegraphics[valign=c,width=0.35\textwidth]{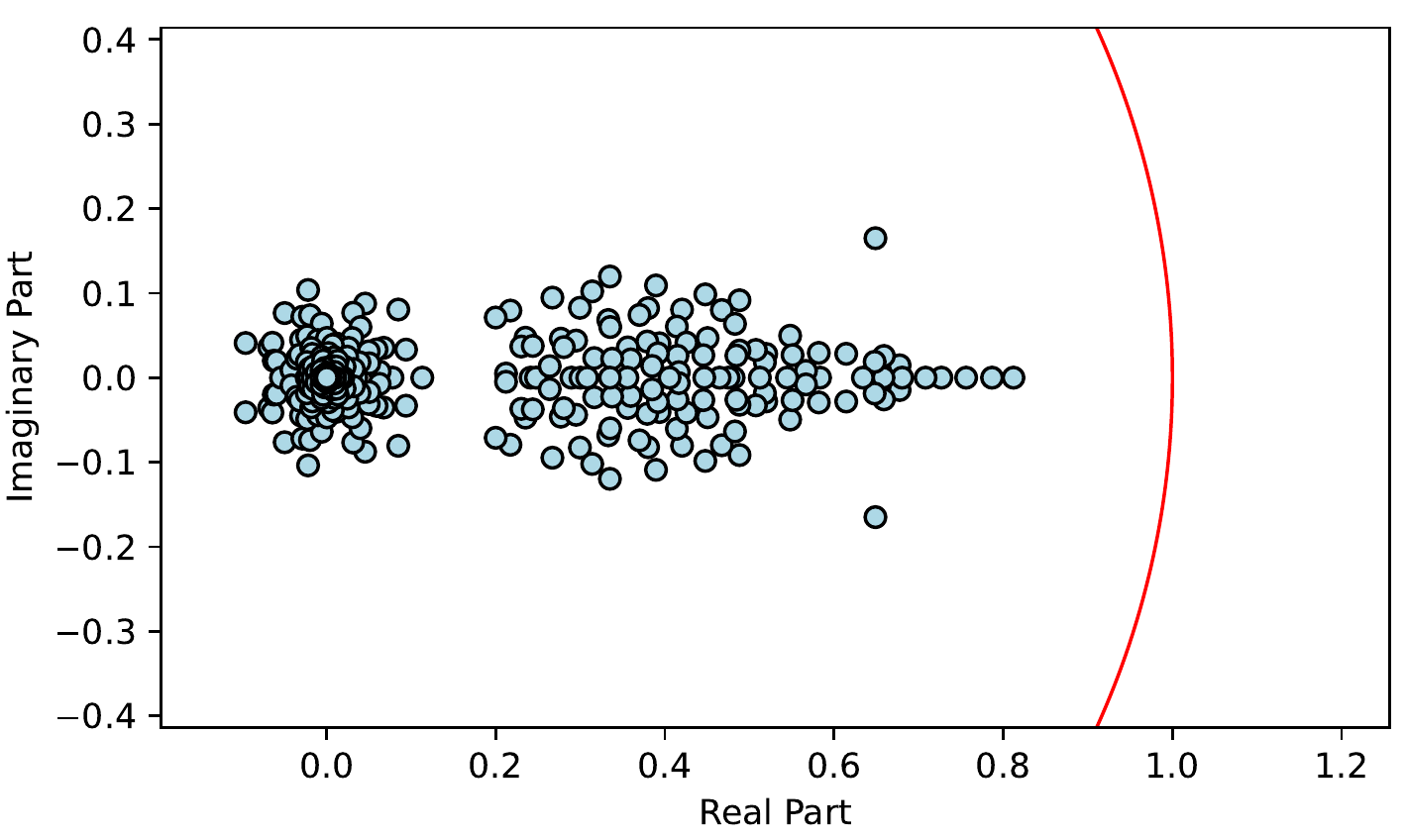}
    \includegraphics[valign=c,width=0.30\textwidth]{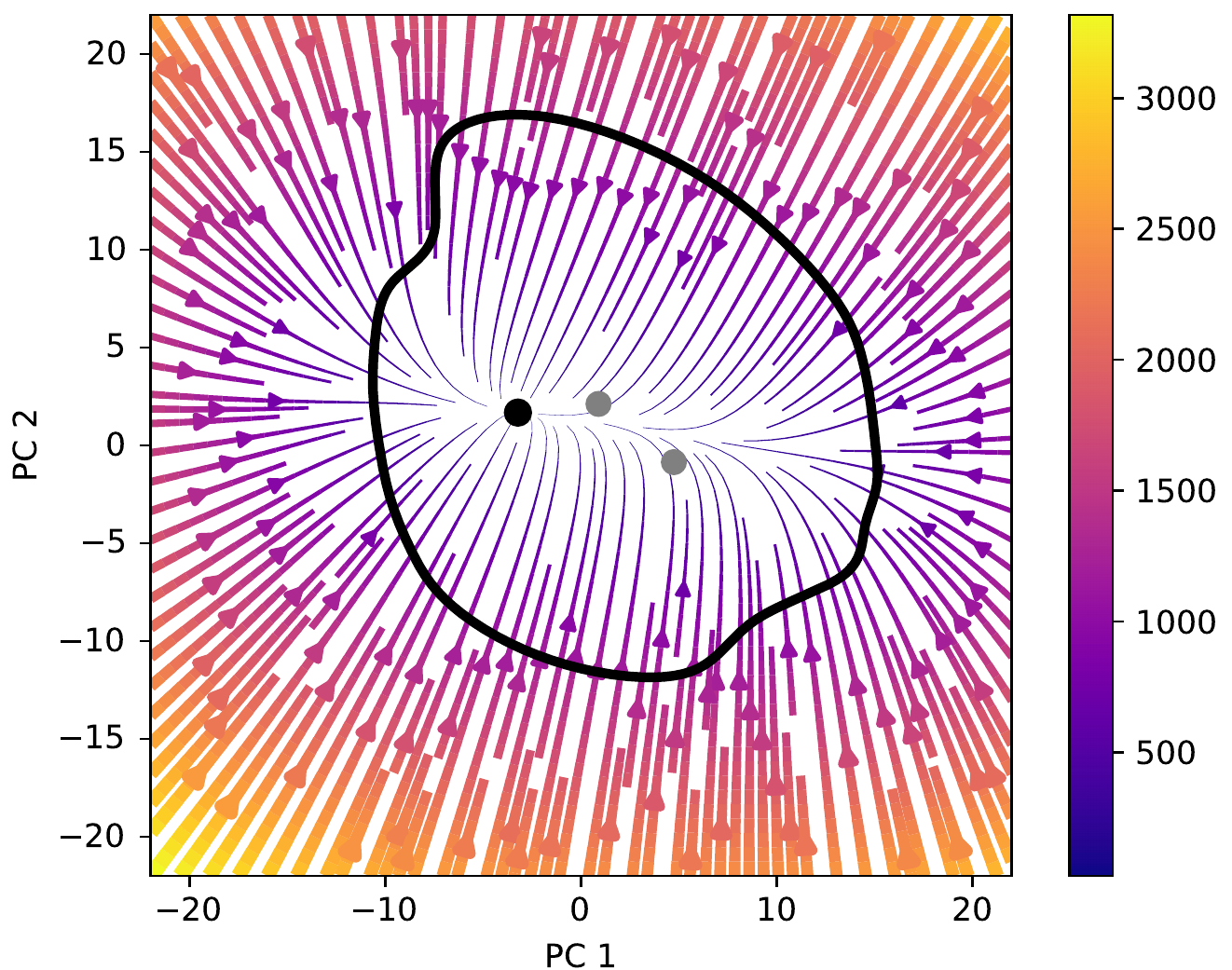}
    \includegraphics[valign=c,width=0.25\textwidth]{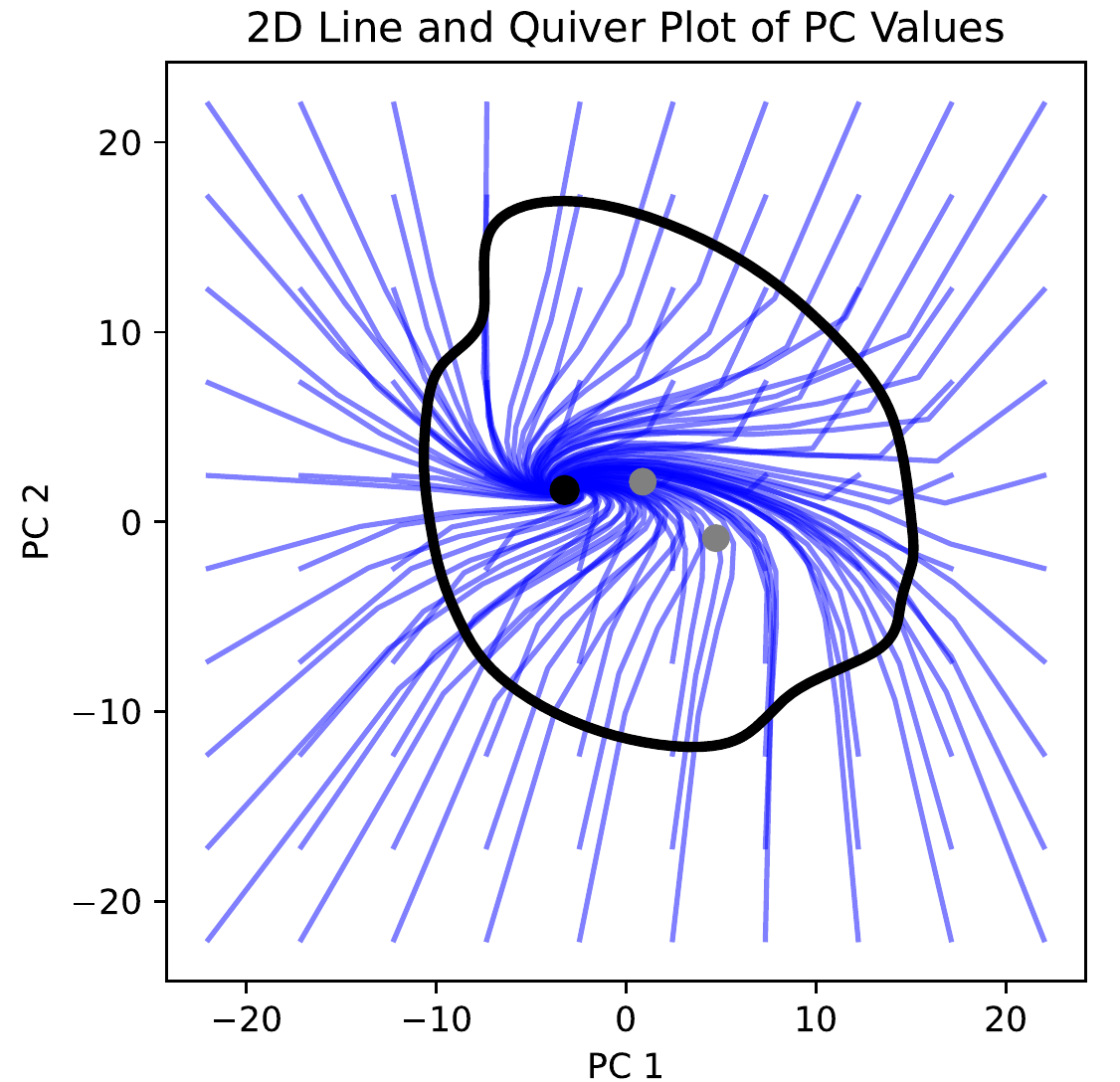}
    
    \includegraphics[valign=c,width=0.35\textwidth]{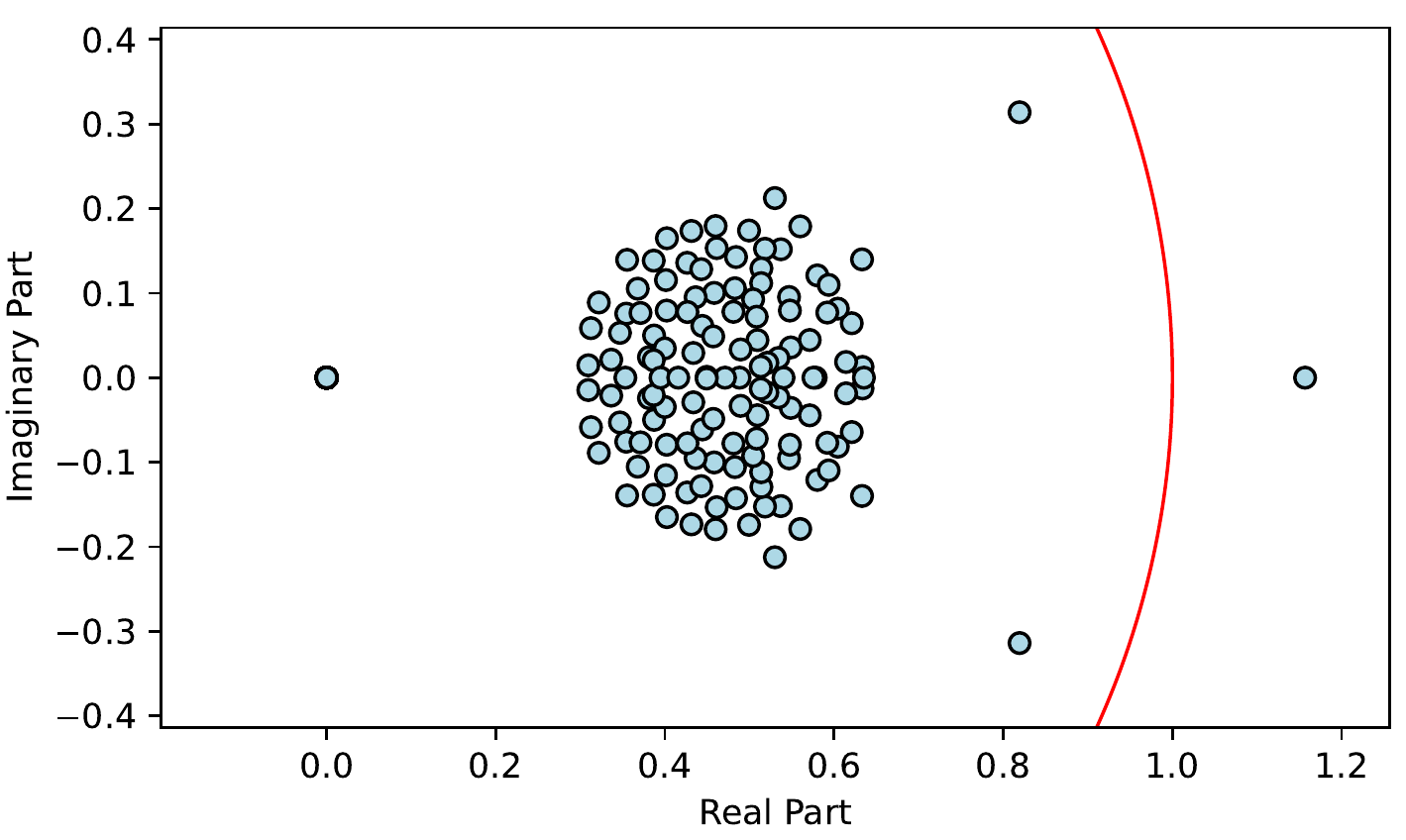}
    \includegraphics[valign=c,width=0.30\textwidth]{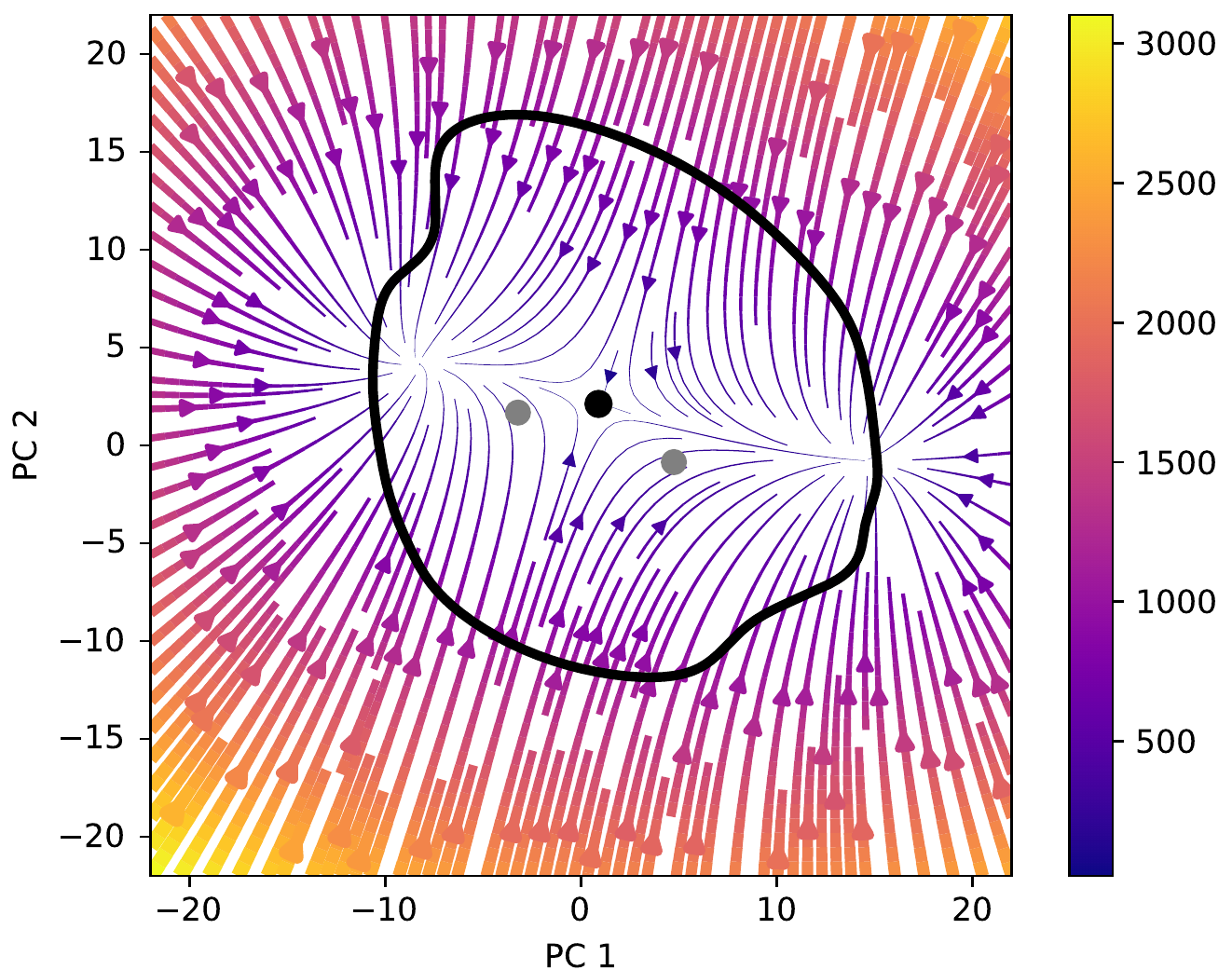}
    \includegraphics[valign=c,width=0.25\textwidth]{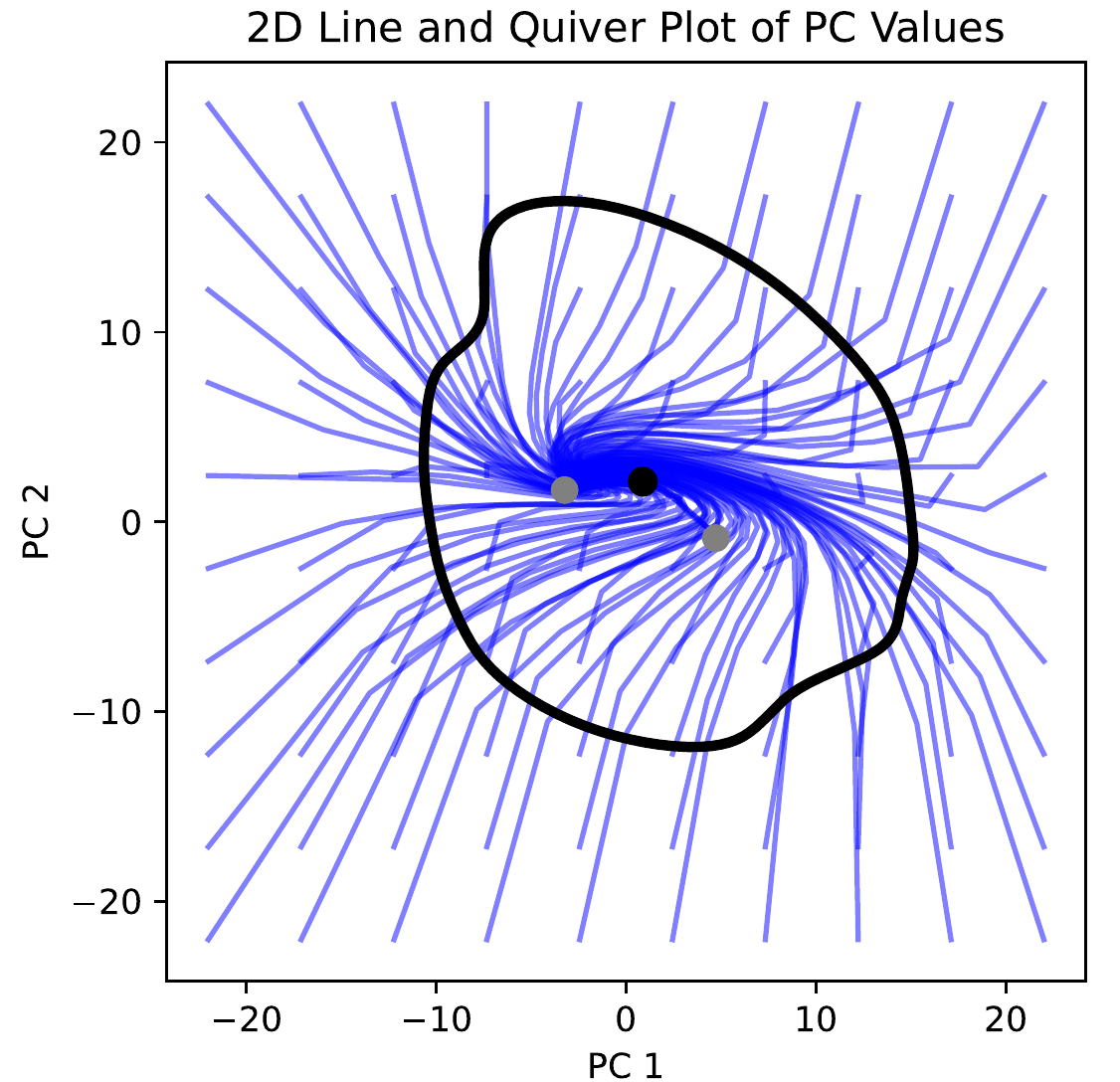}
    
    \subfloat[]{
    \includegraphics[valign=c,width=0.35\textwidth]{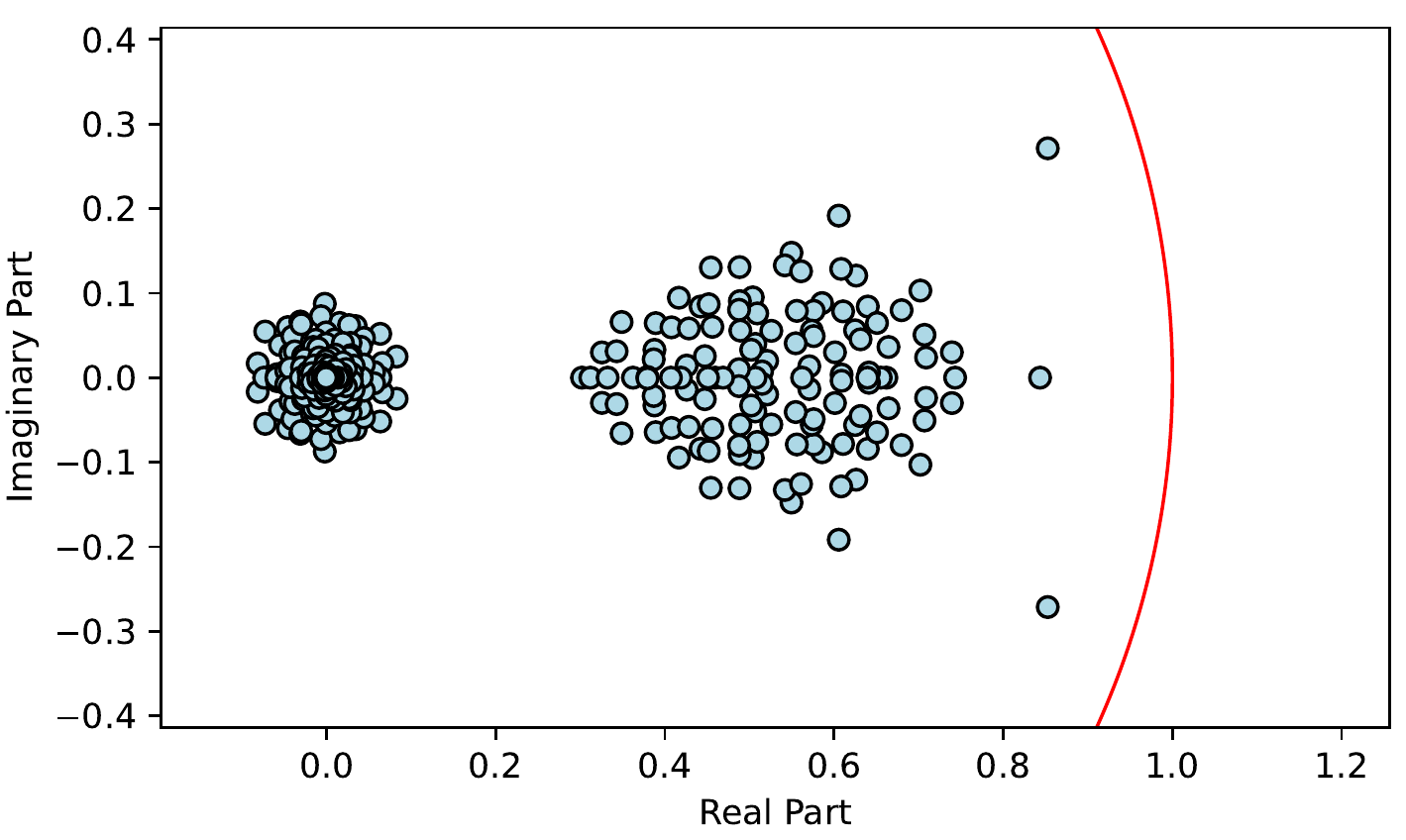}}
    \subfloat[]{
    \includegraphics[valign=c,width=0.30\textwidth]{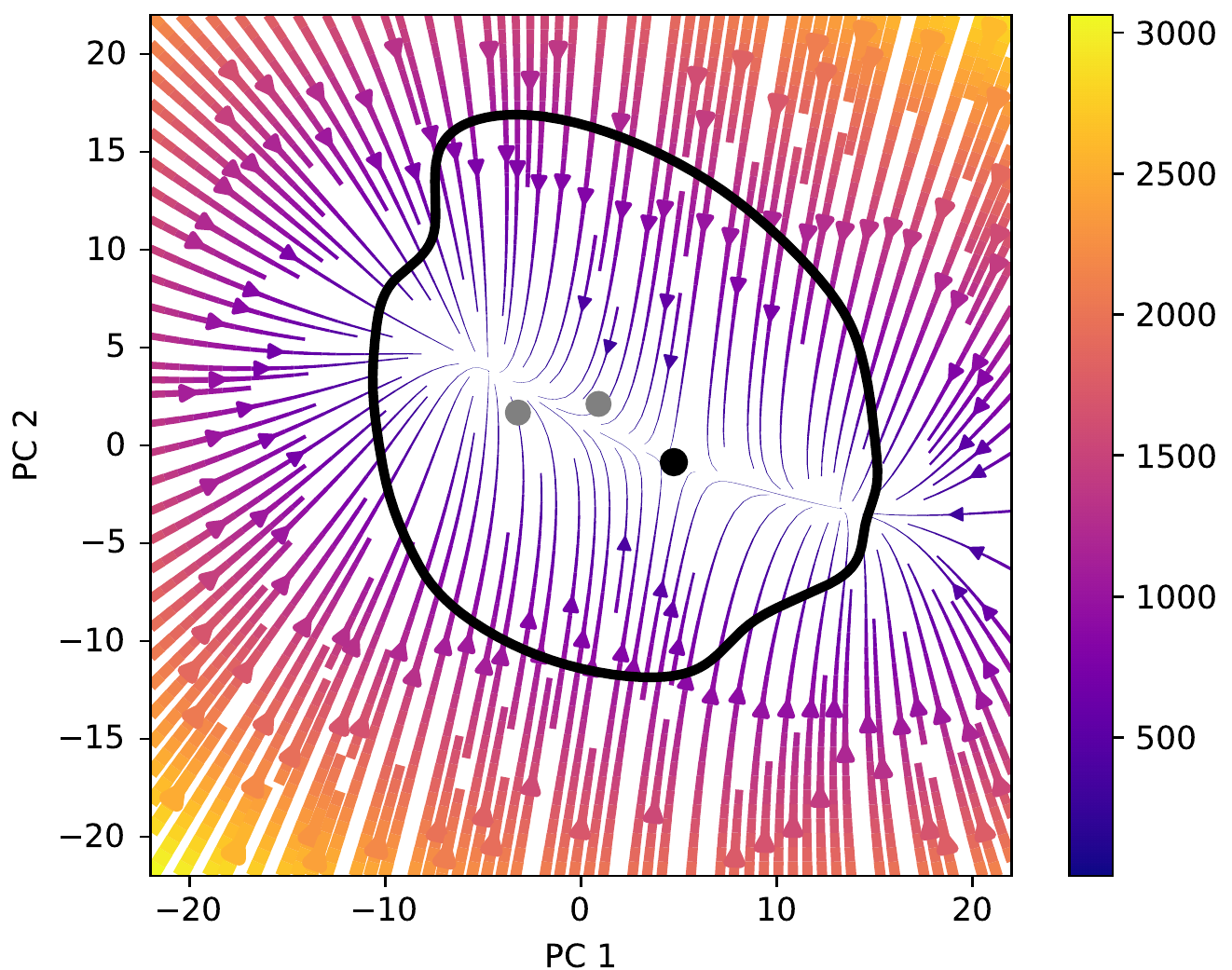}}
    \subfloat[]{
    \includegraphics[valign=c,width=0.25\textwidth]{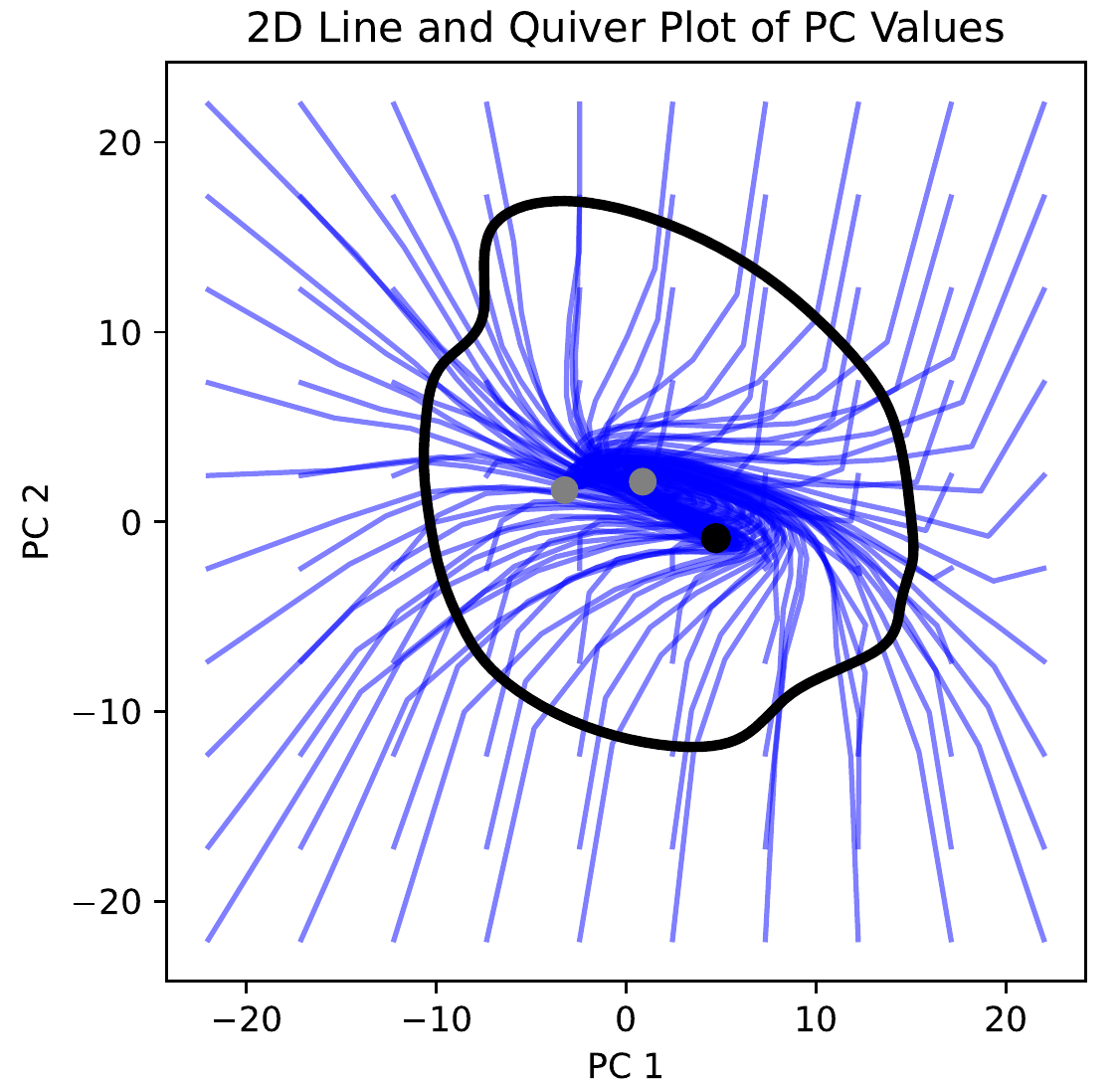}}
    
    \caption{Visualizations of the unforced dynamics of the LSTM16 model, with (left) a catalogue of the eigenvalues associated with each of the three fixed points, (middle) the PC1-PC2 projection of gradients in the system nearby each respective fixed point, and (right) unforced temporal response of the system with recurrent state initial conditions nearby each respcective fixed points.} 
    \label{fig:fixed_points_LSTM16}
\end{figure}

\textbf{Multiple Fixed Points in LSTM Topological Structure.} The LSTM16 model has three fixed points, which reside in the center of its recurrent state trajectories during forward walking, as shown in Figure \ref{fig:3d_LSTM16}. Two are attractors, and the third is a saddle point with one unstable direction, as is reflected in their eigenvalue spectra shown in Figure \ref{fig:fixed_points_LSTM16}(a). The saddle point is located in between two attractors, and its unstable direction points towards them. The gradients centered around each fixed point are projected onto the PC1-PC2 plane in Figure \ref{fig:fixed_points_LSTM16}(b), illustrating the pull of the two attractors and unstable direction of the saddle point. This however, is an oversimplified representation, since the recurrent states do not remain on the PC1-PC2 plane as they decay towards the two stable fixed points. Figure \ref{fig:fixed_points_LSTM16}(c) illustrates the projection of the unforced LSTM dynamics, and how the recurrent states exhibit rotational dynamics as they converge to the two attractors. This rotational trajectory pattern is not obvious from the gradients in Figure \ref{fig:fixed_points_LSTM16}(b), since it assumes recurrent activity and gradients only vary on the PC1-PC2 plane.

\textbf{Robust Agents Have Better Balance.} In contrast, the LSTM4 model is more complicated, as shown in the Supplementary Material, with seven fixed points: two attractors, one marginally stable fixed point, and four saddle points. Geometrically speaking, the two attractors live farther away from the center of the nomimal trajectory, as compared to the LSTM4 attractors. This leads to distinctly different behavior when all sensory input is zero, and the agent is commanded to stand in place. We observe the LSTM4 agents exhibit greater neural activity, which appears in the form of a stepping-in place-motion. In contrast, the neural activity of the LSTM16 networks is relatively static, which enables the agent to balance in place without stepping. Note that because the MLP network includes bias terms, having a sensory input of zero reduces, but does not completely zero out the input to the LSTM networks. If agents were trained with zero MLP bias, we would expect to see both LSTM4 and LSTM16 neural activity to converge to an attractor and result in the robot standing still.




\subsection{Neural Perturbations}


\textbf{Evidence of Structured Low-Dimensional Neural Dynamics.} We apply targeted neural perturbations along the top PC directions, and identify two distinct patterns. First, we find that perturbing neurons along the dominant PC directions elicits a larger magnitude response, and also implicate other neural populations as well. Perturbed responses converge back to nominal activity within less than a half of a gait cycle. Second, neural perturbations cause a phase shift in the gait cycle when perturbations are tangential to the instantaneous direction of neural activity, effectively interacting with the neural dynamics. However, when perturbations are orthogonal to the direction of neural activity, the system is virtually unaffected. Together, these two findings support the hypothesis that our RNN-based controller exhibits structured low-dimensional neural dynamics, similar to findings in primates \cite{oshea_direct_2022}.

\begin{figure}[h]
    \centering
    \includegraphics[width=0.65\textwidth]{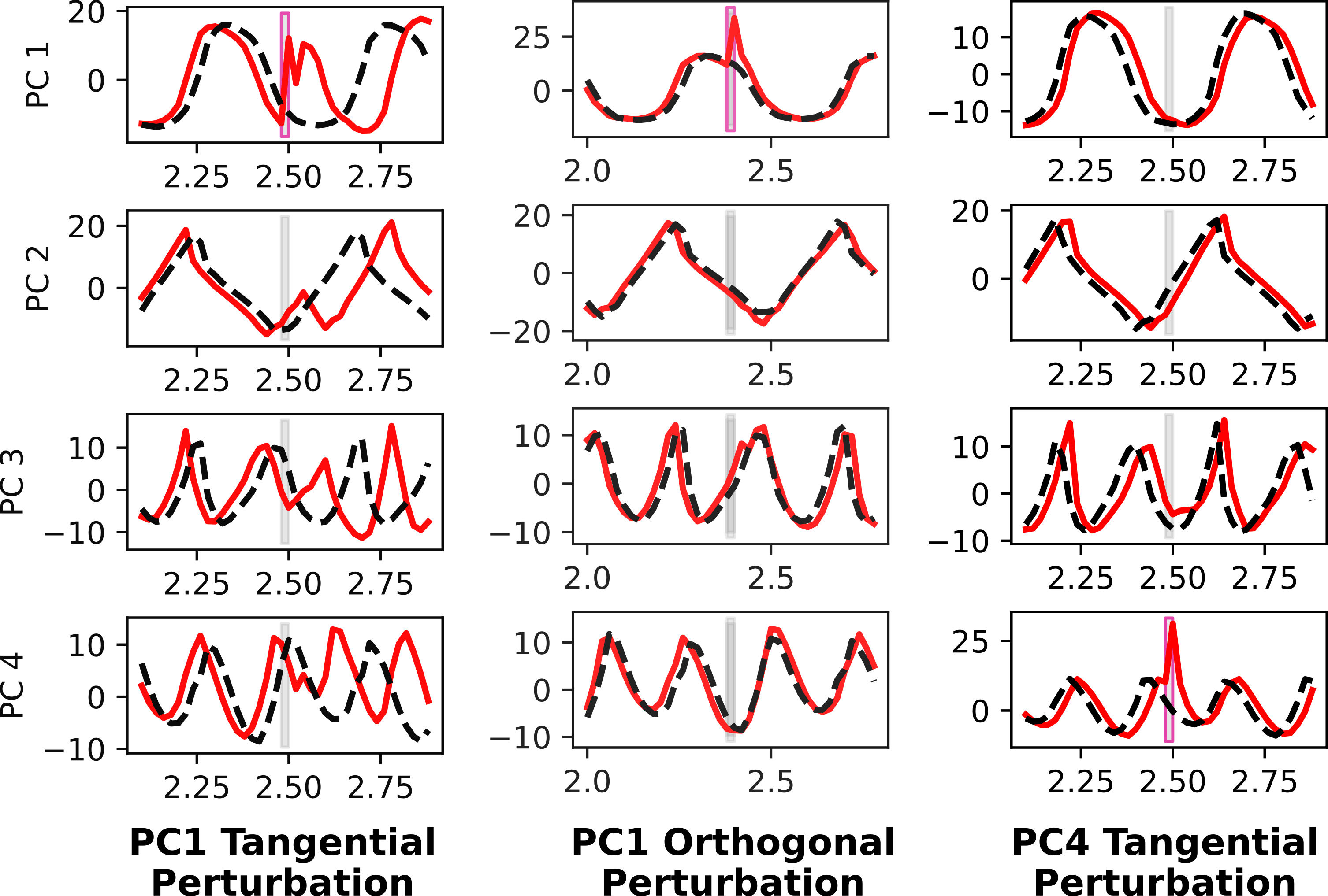}
    \caption{Response of recurrent states in PC1 through PC4 directions, before and after targeted neural perturbations. A perturbation is applied to PC1 that is tangential to the PC1-PC2 plane (left), which causes a significant phase shift in the gait cycle. This is evident in PC1 through PC4. However, the same perturbation applied at later time step is orthogonal to the PC1-PC2 plane (middle), and doesn't cause any disruption to the gait cycle. When applying a perturbation in the PC4 direction, there is little impact on the population activity, despite it being tangential to the PC1-PC4 plane.}
    \label{fig:neural_hc_pc_perturb_LSTM4}
\end{figure}




\subsection{Physical Perturbations}

As a practical matter, robustness to physical disturbances is a defining characteristic of animals, and more recently, state-of-the-art legged robots. As discussed in Section \ref{sec:method} section, we find that the LSTM16 model trained with a BPTT truncation length of 16 is more robust to disturbances.

\begin{figure}[h]
    \centering
    \includegraphics[width=0.9\textwidth]{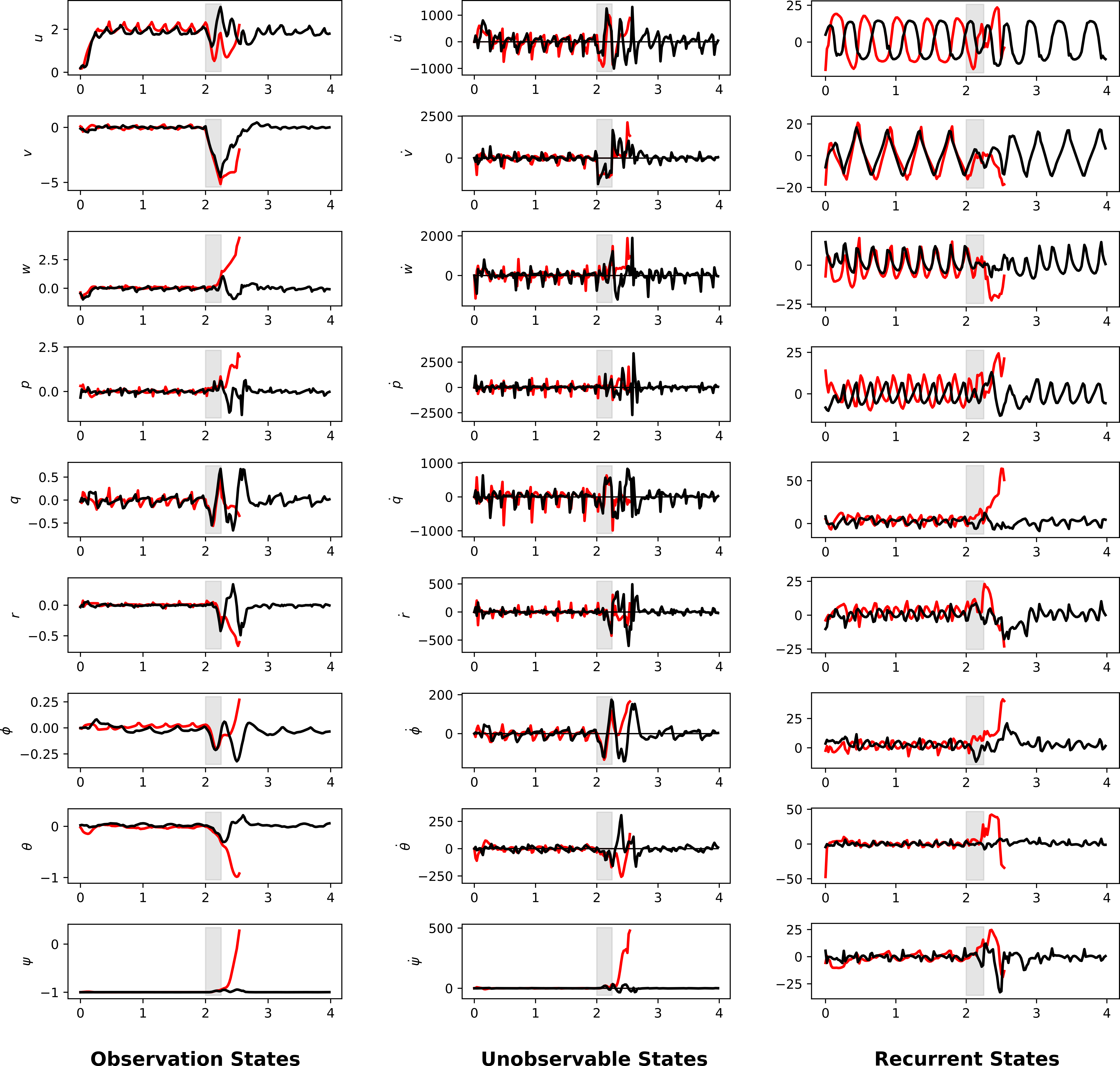}
    \caption{Comparison of RNN-based quadrupedal locomotion controllers, trained with a BPTT sequence length of (red) 4 and (black) 16. We apply a external force laterally for a direction of 250ms (shaded grey region), as the robot is walking forward at a speed of $2$ m/s. We find that the body velocities in the observation vector (left) do not capture the full state of the agent, since external forces result in (middle) body accelerations which are unobservable. In particular, we see lateral body acceleration $\dot{w}$ and body angular accelerations $\dot{r}$ experience persistent offsets for the duration of the perturbation. We see significant deviation in the (right) recurrent states relative to nominal operation. } %
    \label{fig:physical_perturb}
\end{figure}

We compare single LSTM16 and LSTM4 trials that display stereotypical behavior, and find that the body accelerations $\dot{v}$ and $\dot{p}$ are relatively preserved during the perturbation duration, as shown in Figure \ref{fig:physical_perturb}. These unobserved states may be encoded into the LSTM network. For LSTM4, the truncation window is 80ms, whereas it is 320ms for LSTM16. Since this experiment is done with 100ms perturbations, the LSTM16 model may be more capable of encoding these accelerations for the full duration of the perturbation.

Additionally, the longer BPTT truncation length of LSTM16 may enable a more complex spatiotemporal response that aids in recovering from disturbances. For larger magnitude disturbances, it generally takes longer to recover and return to normal walking.




\section{Conclusion}
\label{sec:conclusion}

We leverage a suite of neuroscience methodologies from the computation through dynamics community to interpret and begin to reverse-engineer RNN-based RL controllers. This approach has been powerful for understanding the neural dynamics of robust sensorimotor systems, from identifying marked differences in fixed point topology, to identifying structured, low-dimensional neural behavior. Ideally, future research in this domain would build off this work and enable more robust design principles or modeling solutions.


\textbf{Limitations.}

Regarding the ideas mentioned in the physical perturbation section, additional decoding analysis is required to strengthen those claims. Further efforts to optimize curriculum learning would be beneficial to training of the agent. For this work, the distribution of perturbation magnitudes is constant across training, which is likely not optimal. Additionally further analysis of perturbations at different parts of the gait cycle would be insightful, since agent behavior and robustness may vary. 



\clearpage


\bibliography{references}  

\newpage

\section{Supplementary Materials}

\begin{figure}[h!]
    \centering
    \includegraphics[valign=c,width=0.35\textwidth]{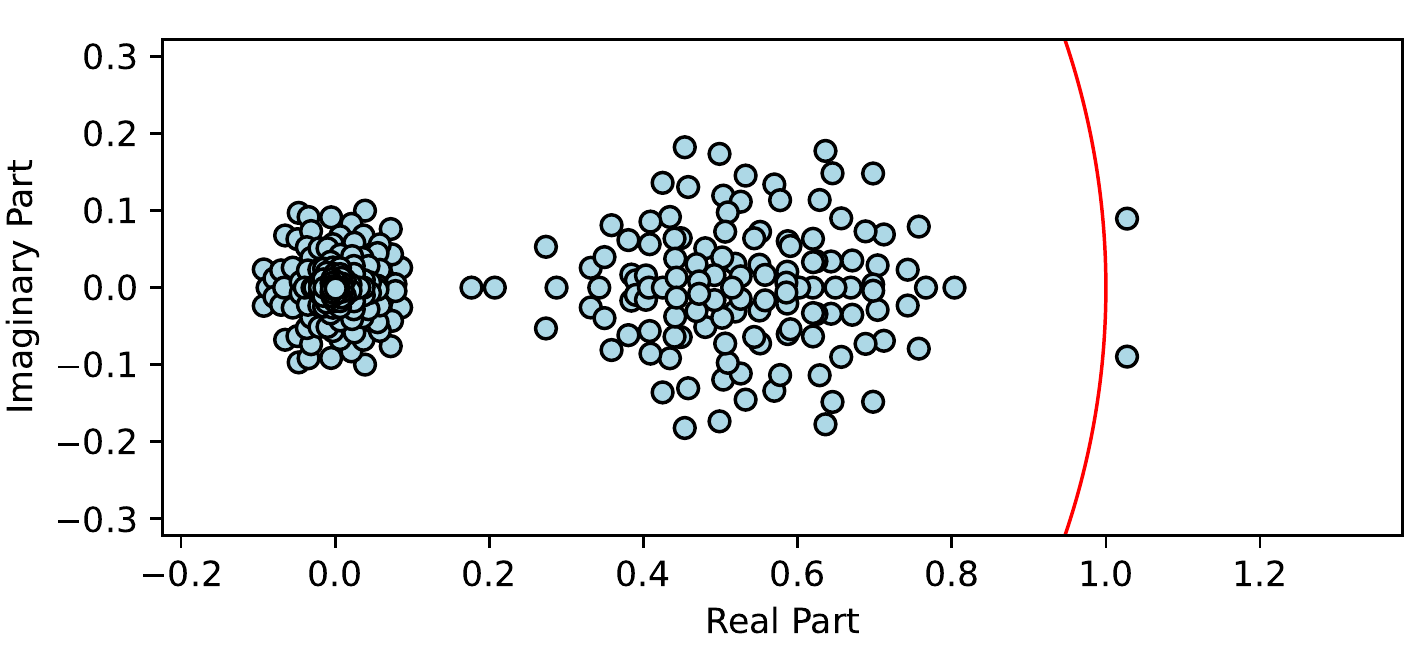}
    \includegraphics[valign=c,width=0.225\textwidth]{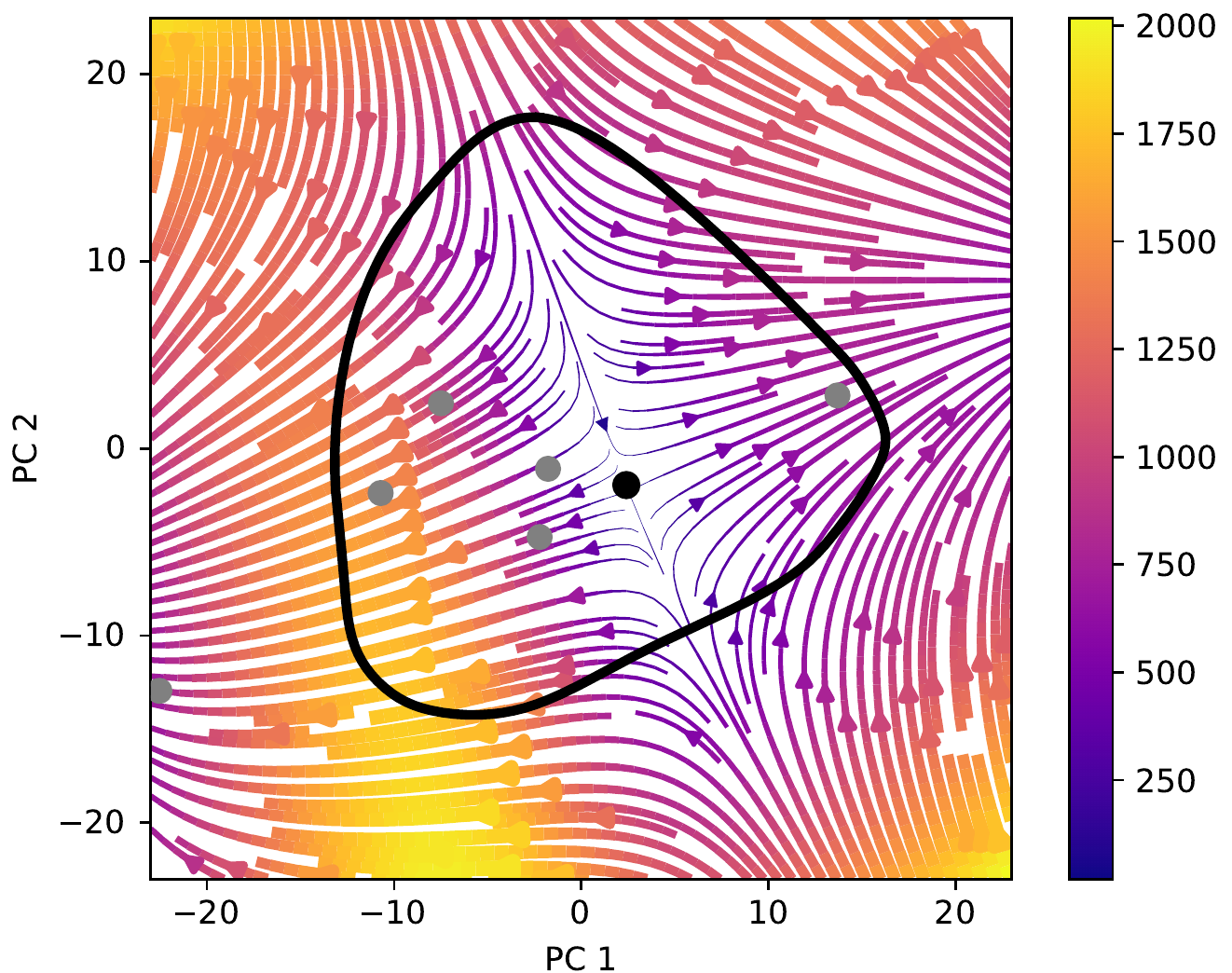}
    \includegraphics[valign=c,width=0.225\textwidth]{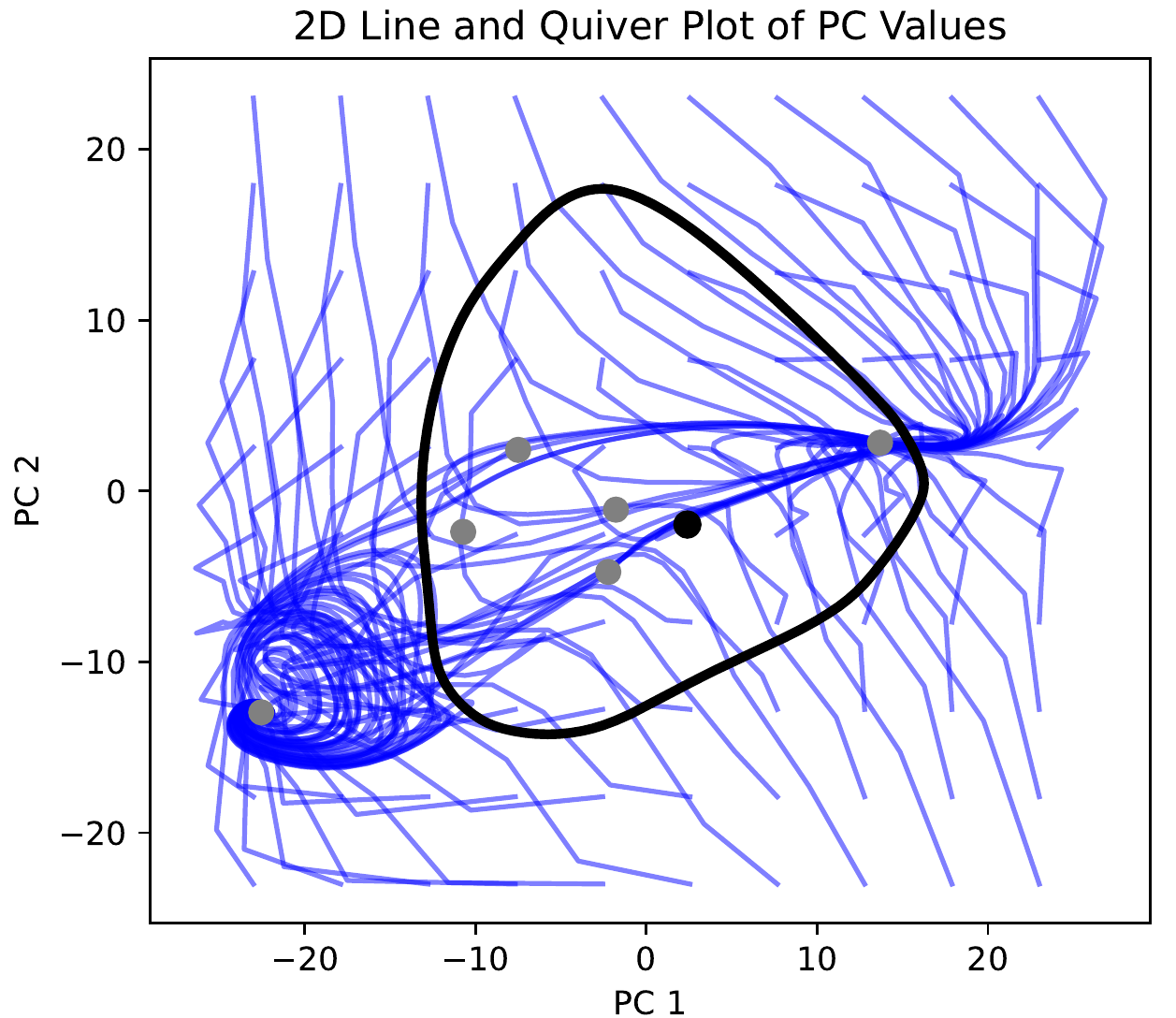} \\
    
    \includegraphics[valign=c,width=0.35\textwidth]{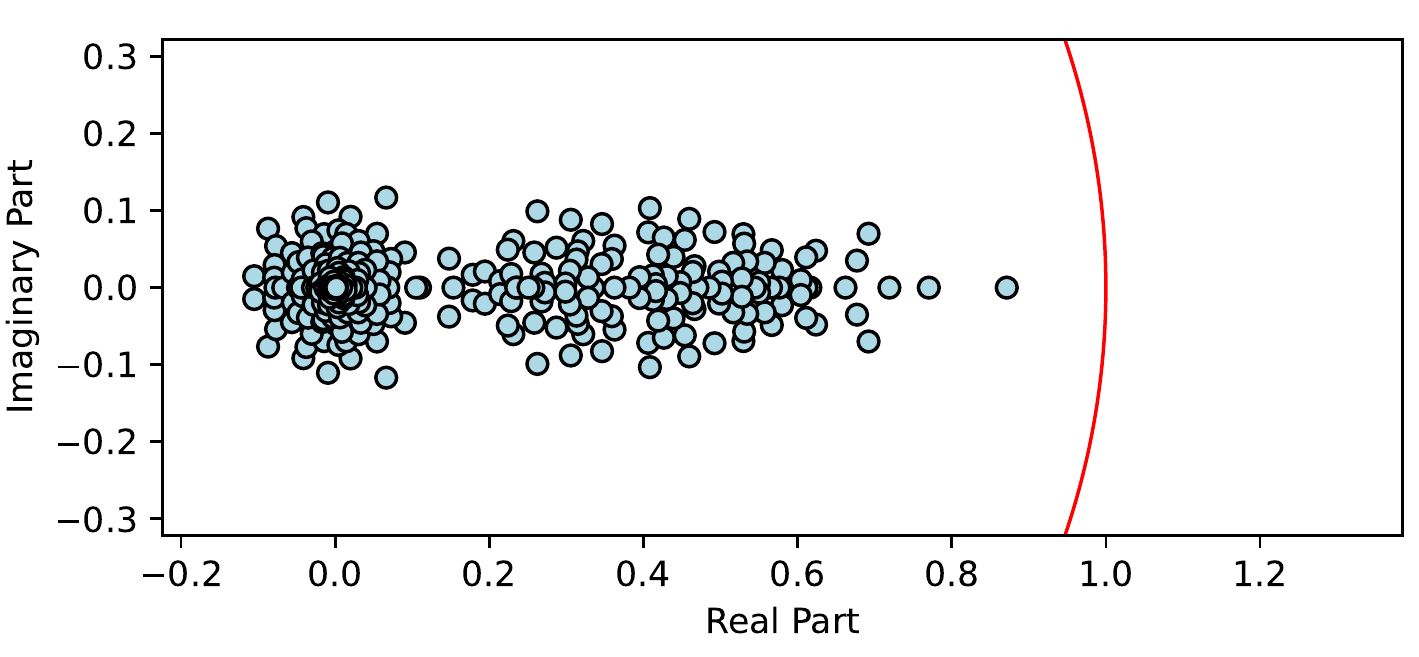}
    \includegraphics[valign=c,width=0.225\textwidth]{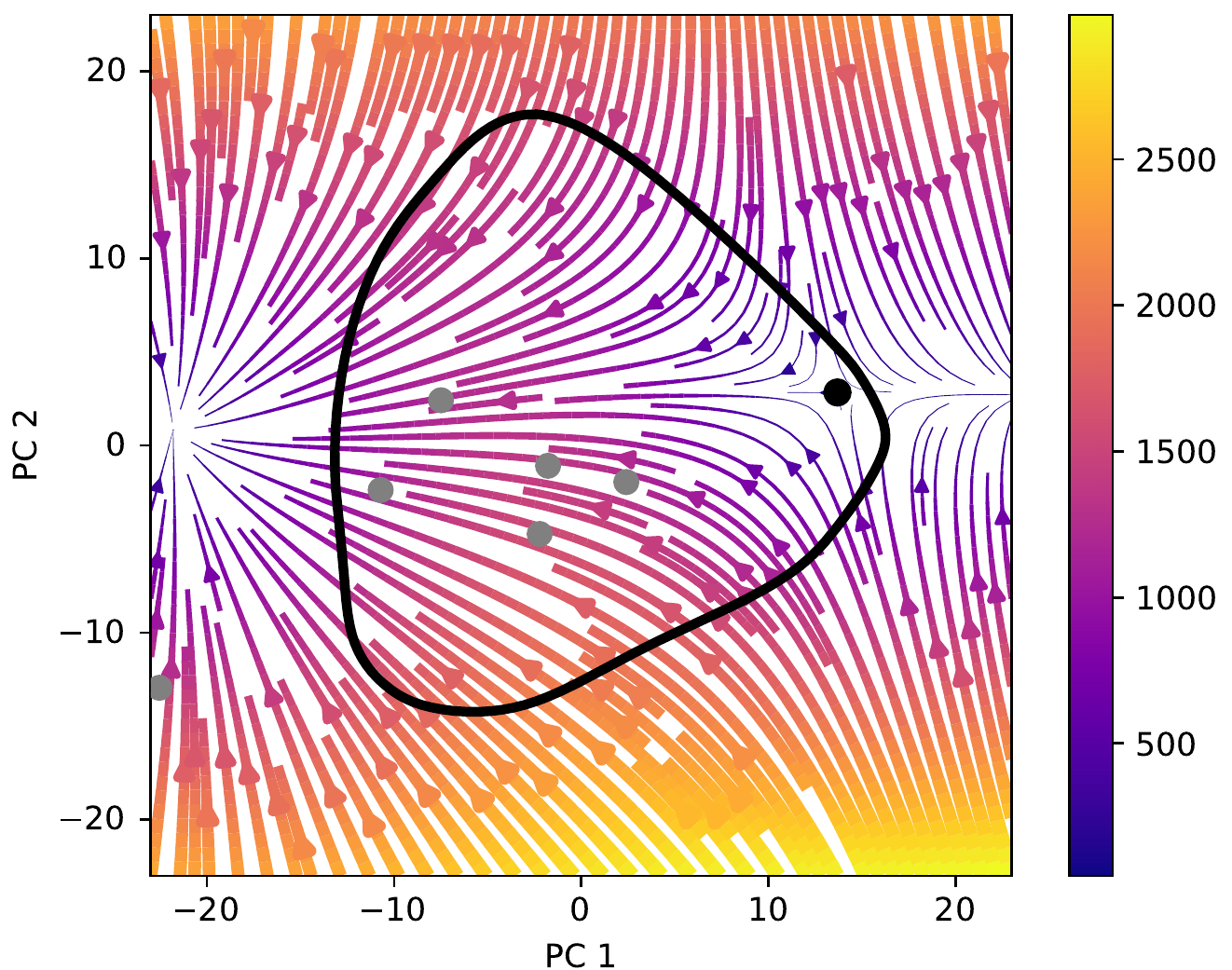}
    \includegraphics[valign=c,width=0.225\textwidth]{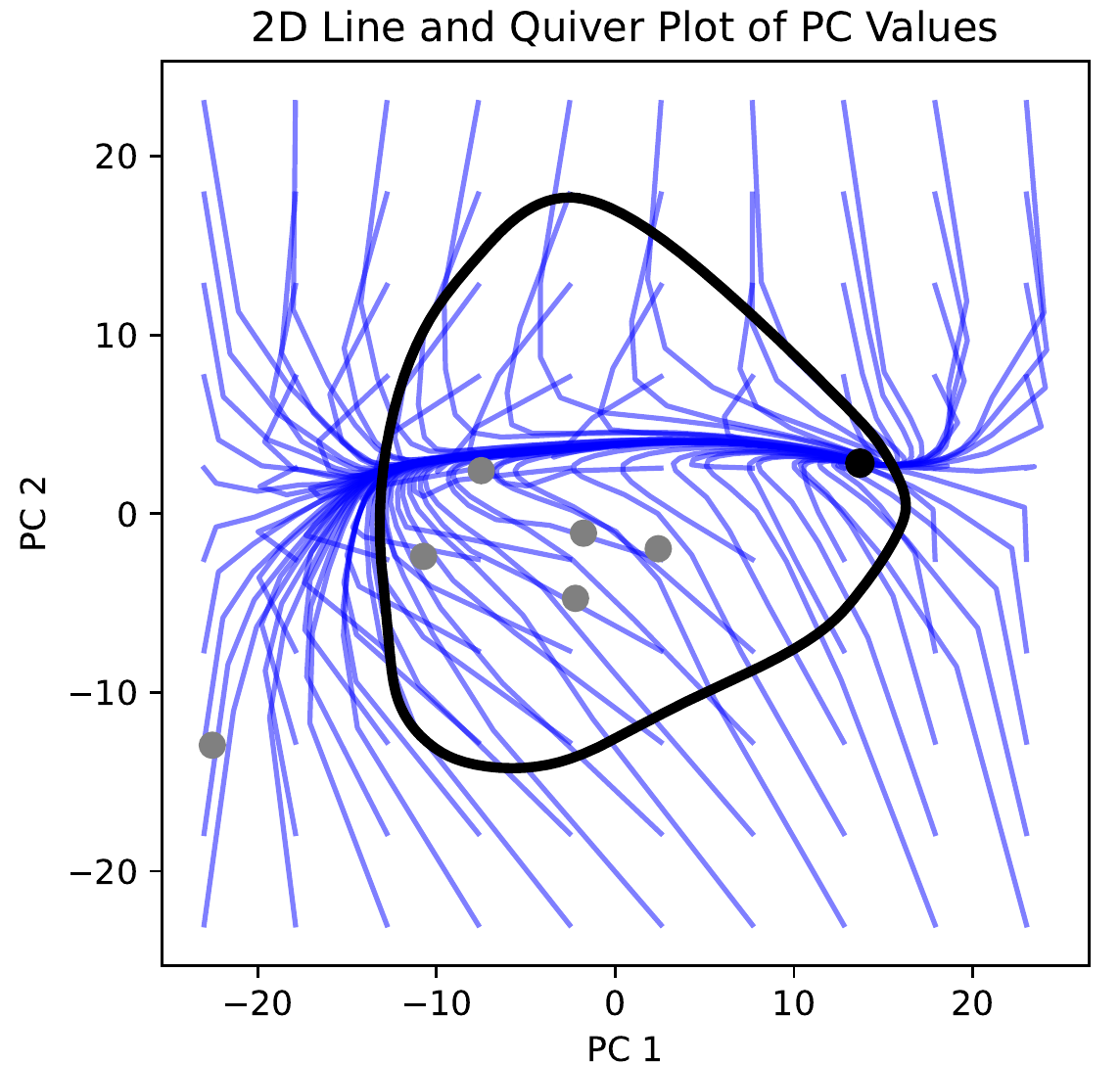} \\
    
    \includegraphics[valign=c,width=0.35\textwidth]{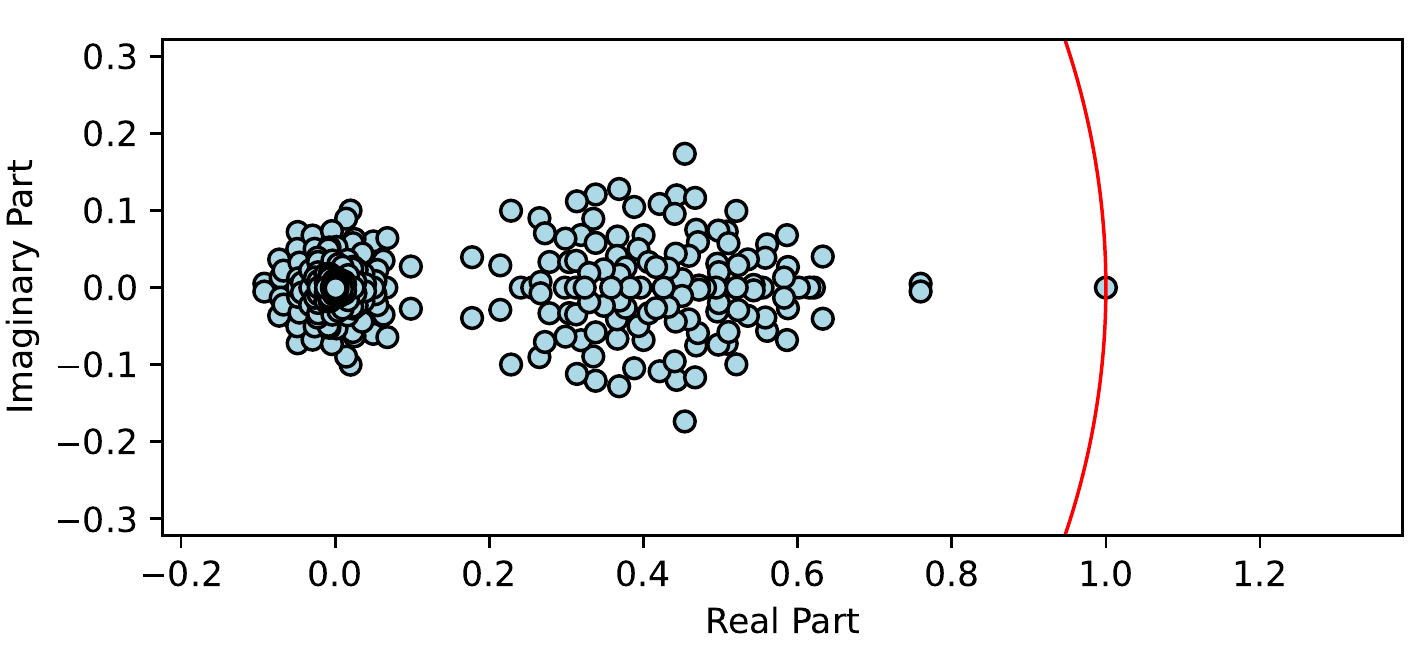}
    \includegraphics[valign=c,width=0.225\textwidth]{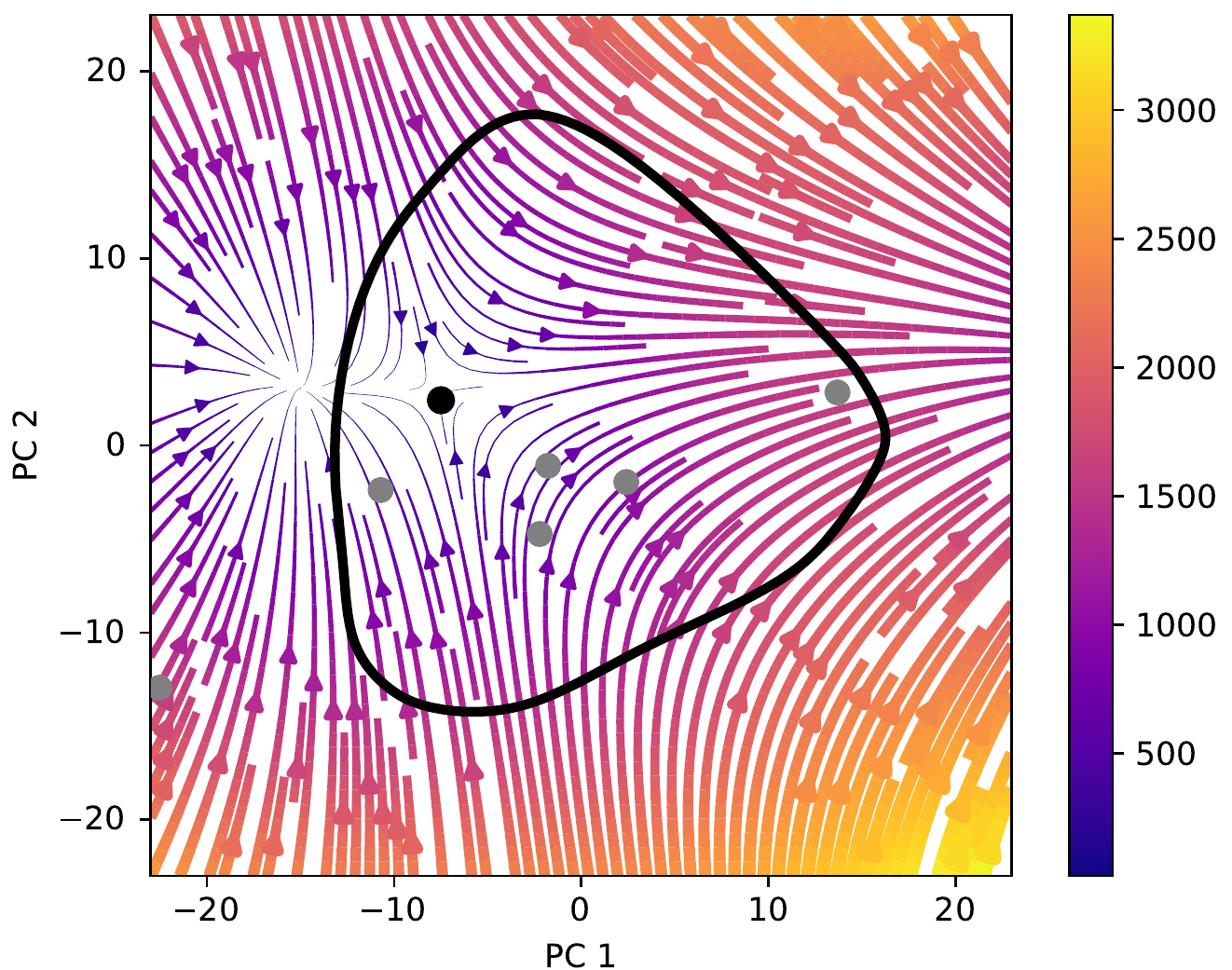}
    \includegraphics[valign=c,width=0.225\textwidth]{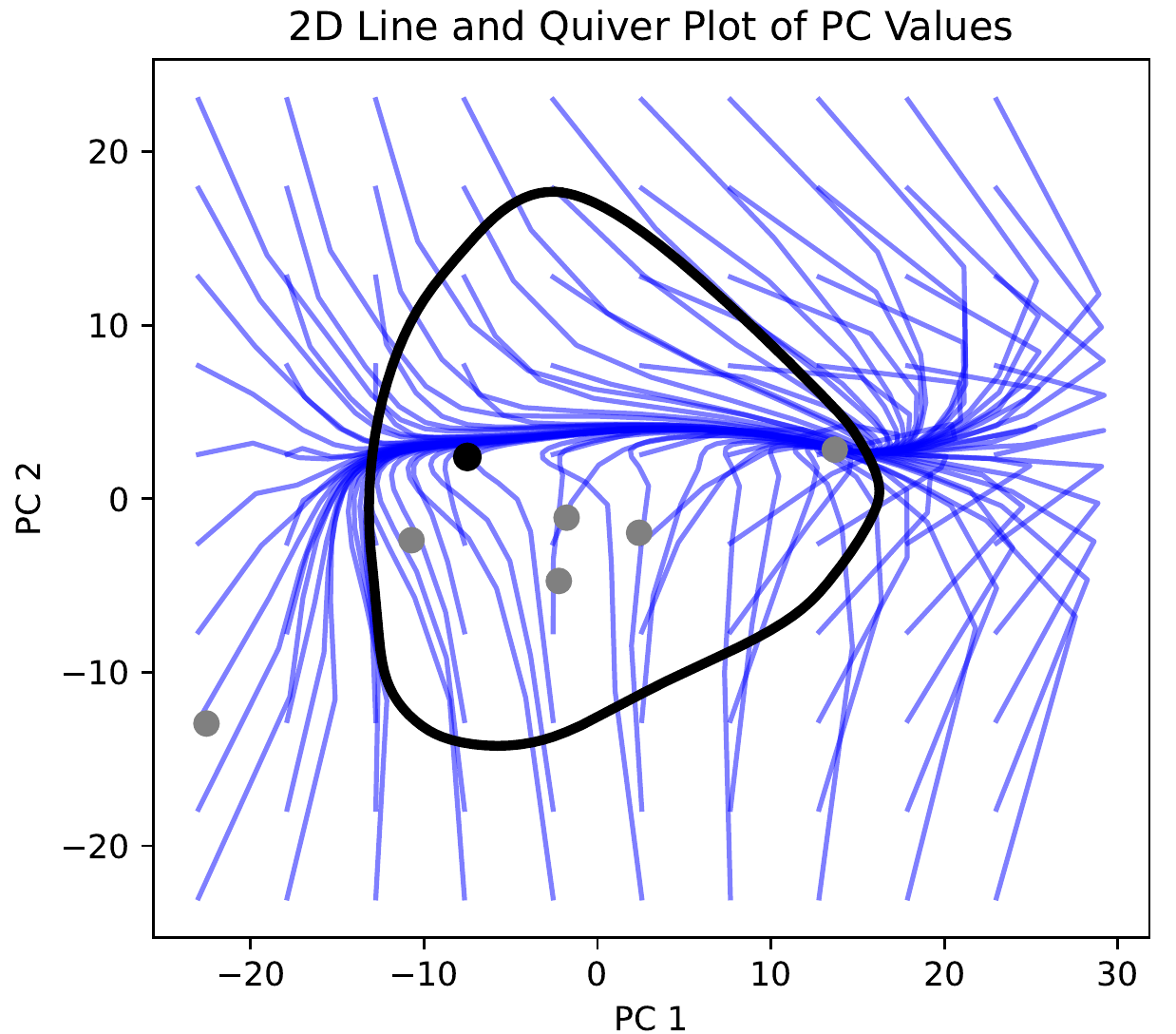} \\
    
    \includegraphics[valign=c,width=0.35\textwidth]{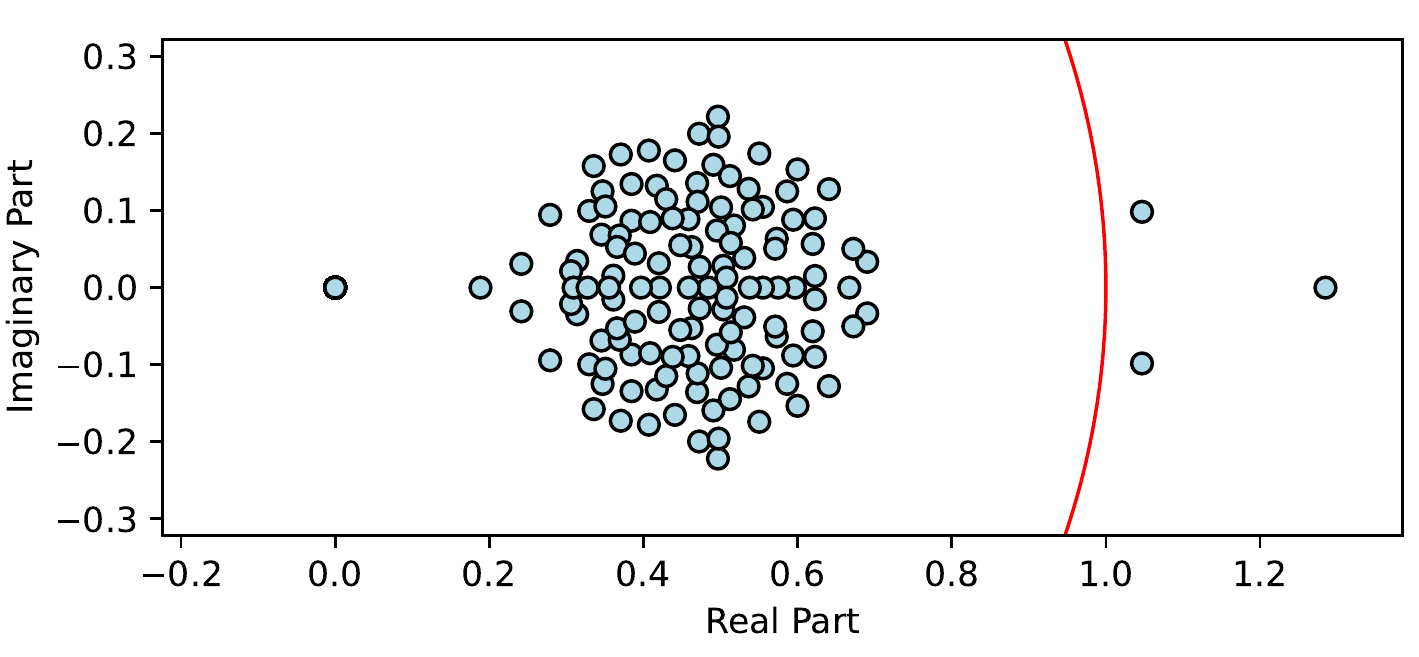}
    \includegraphics[valign=c,width=0.225\textwidth]{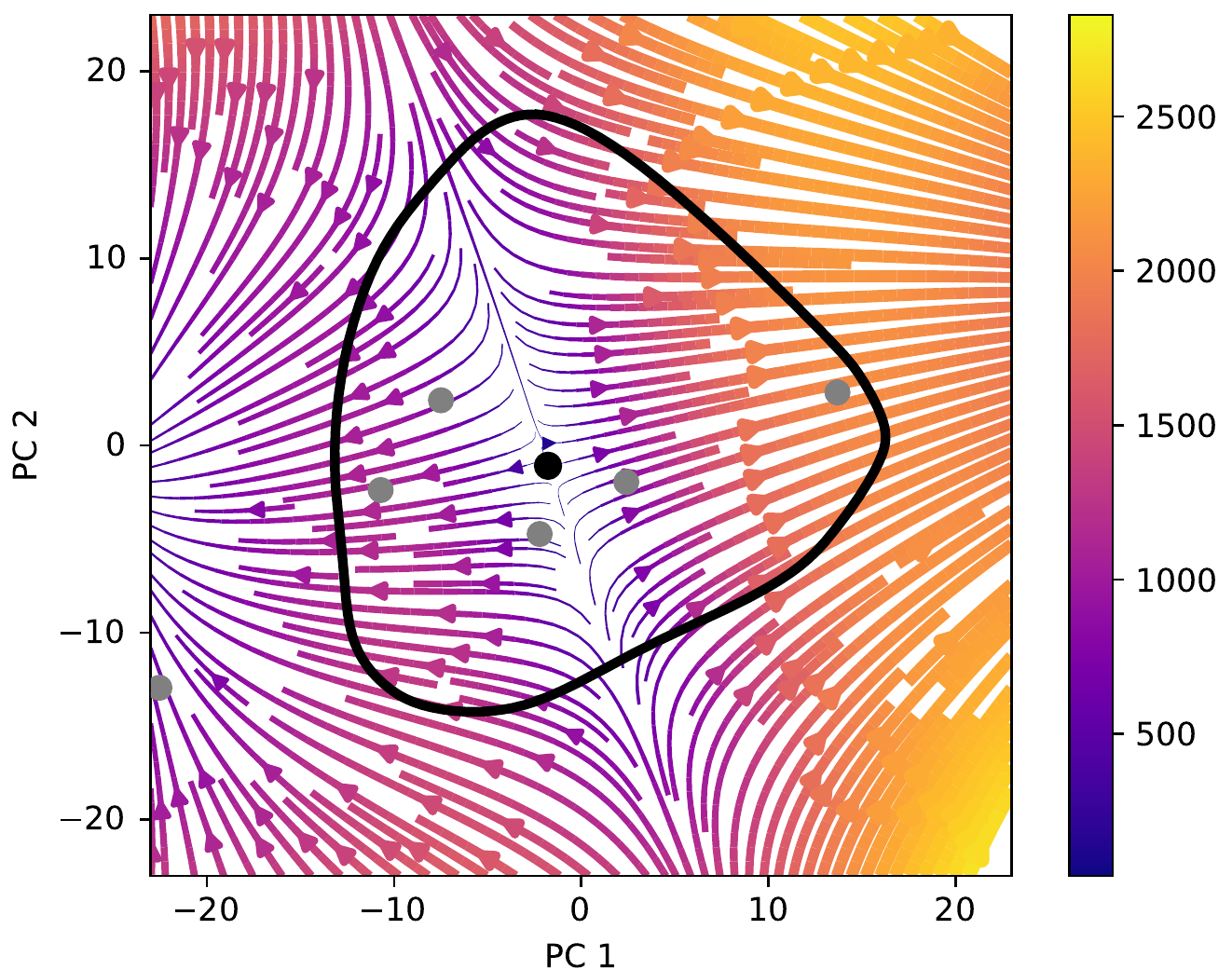}
    \includegraphics[valign=c,width=0.225\textwidth]{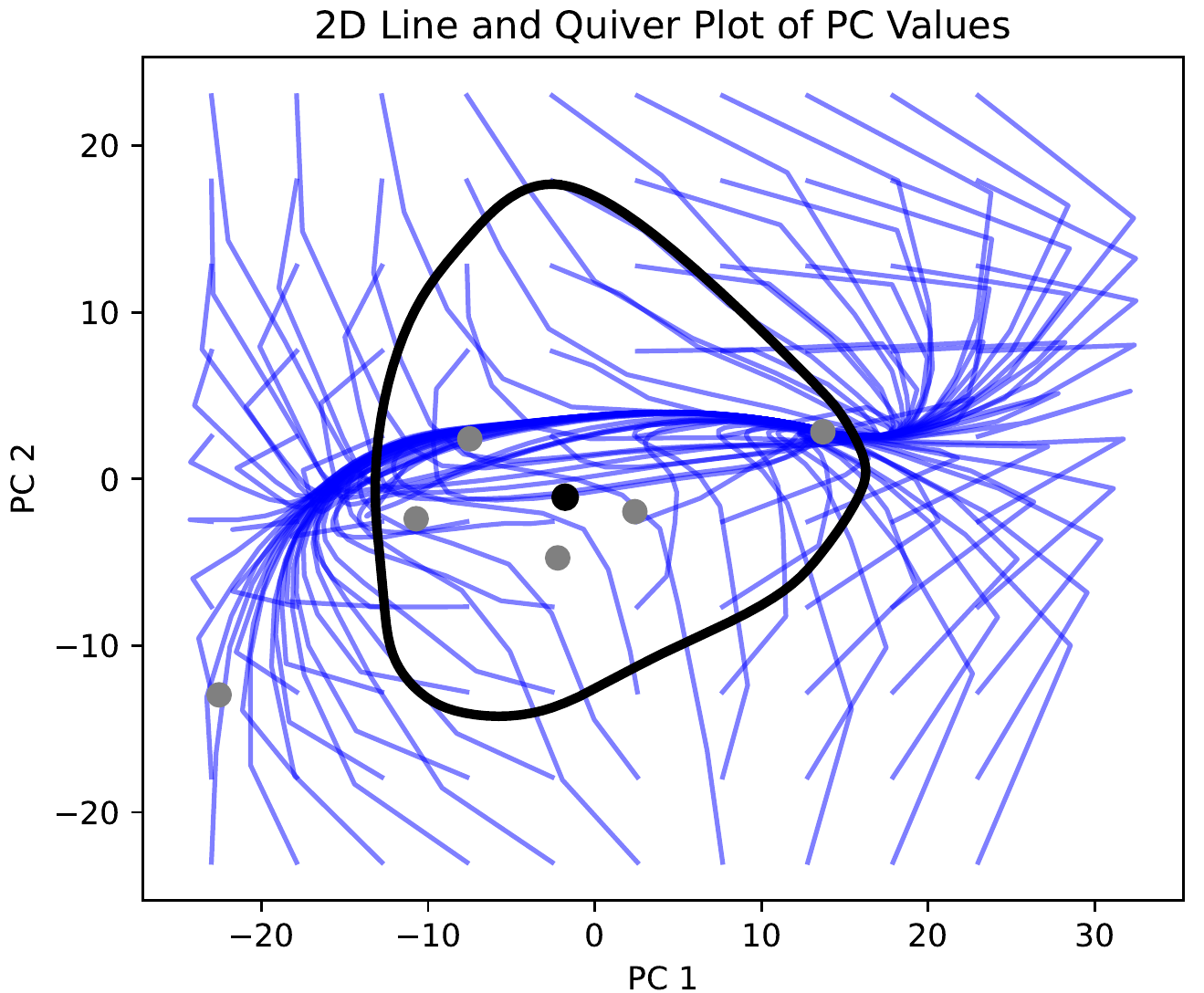} \\
    
    \includegraphics[valign=c,width=0.35\textwidth]{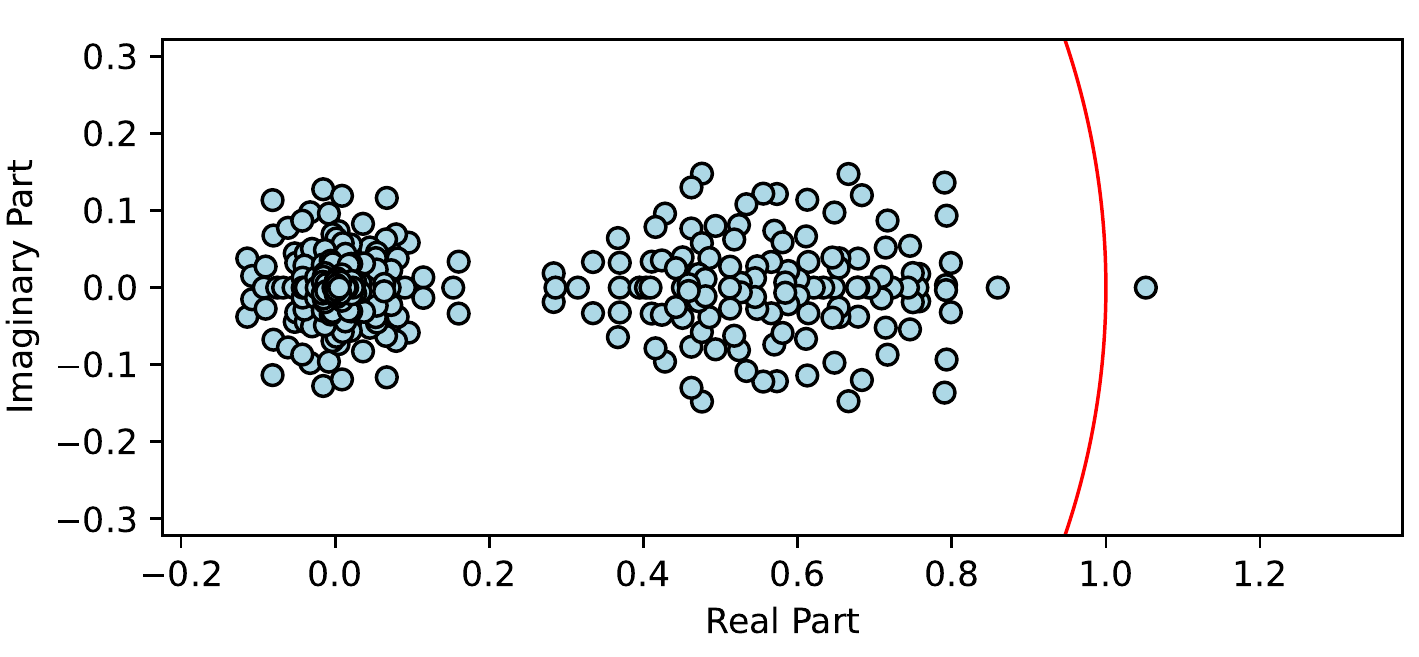}
    \includegraphics[valign=c,width=0.225\textwidth]{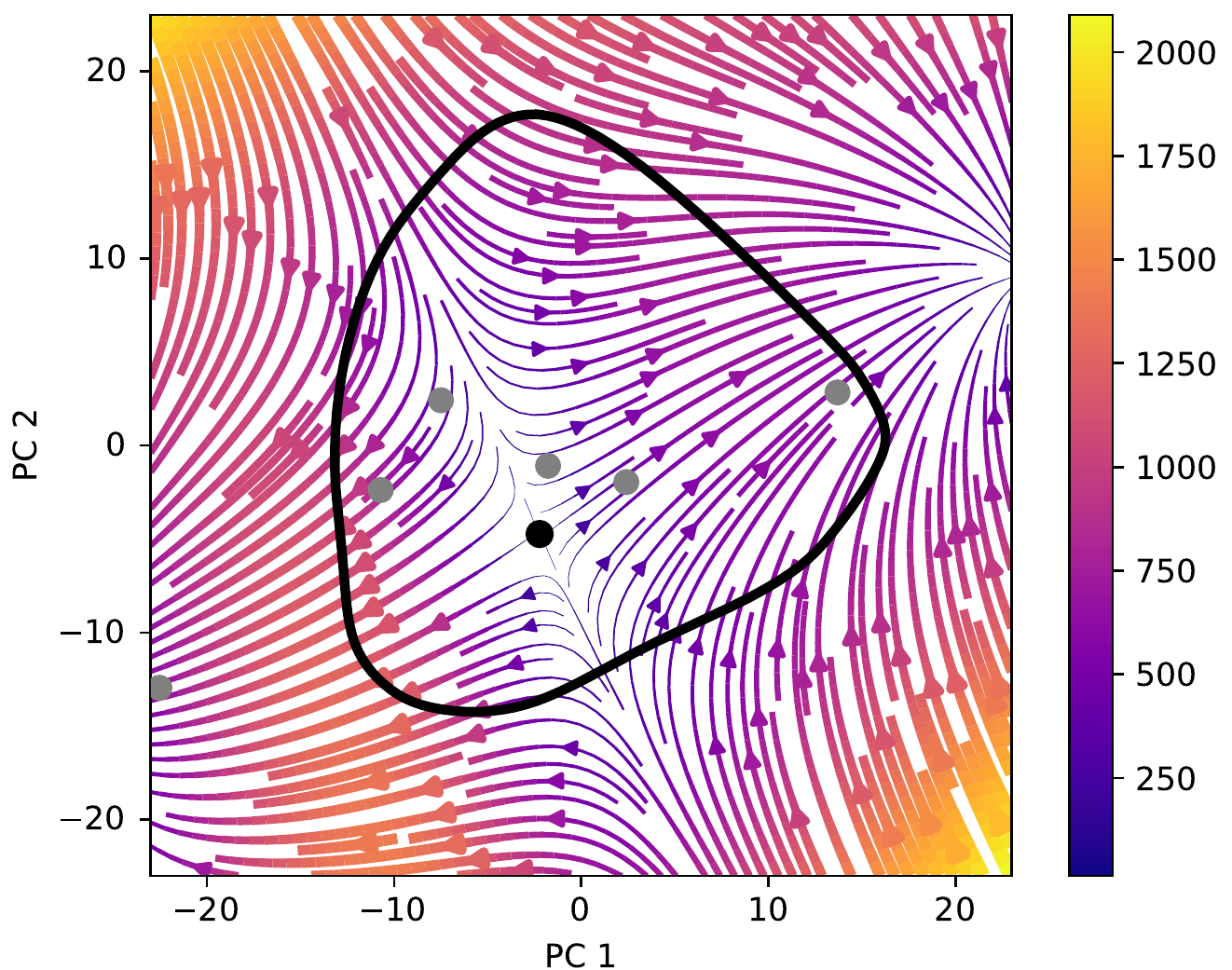}
    \includegraphics[valign=c,width=0.225\textwidth]{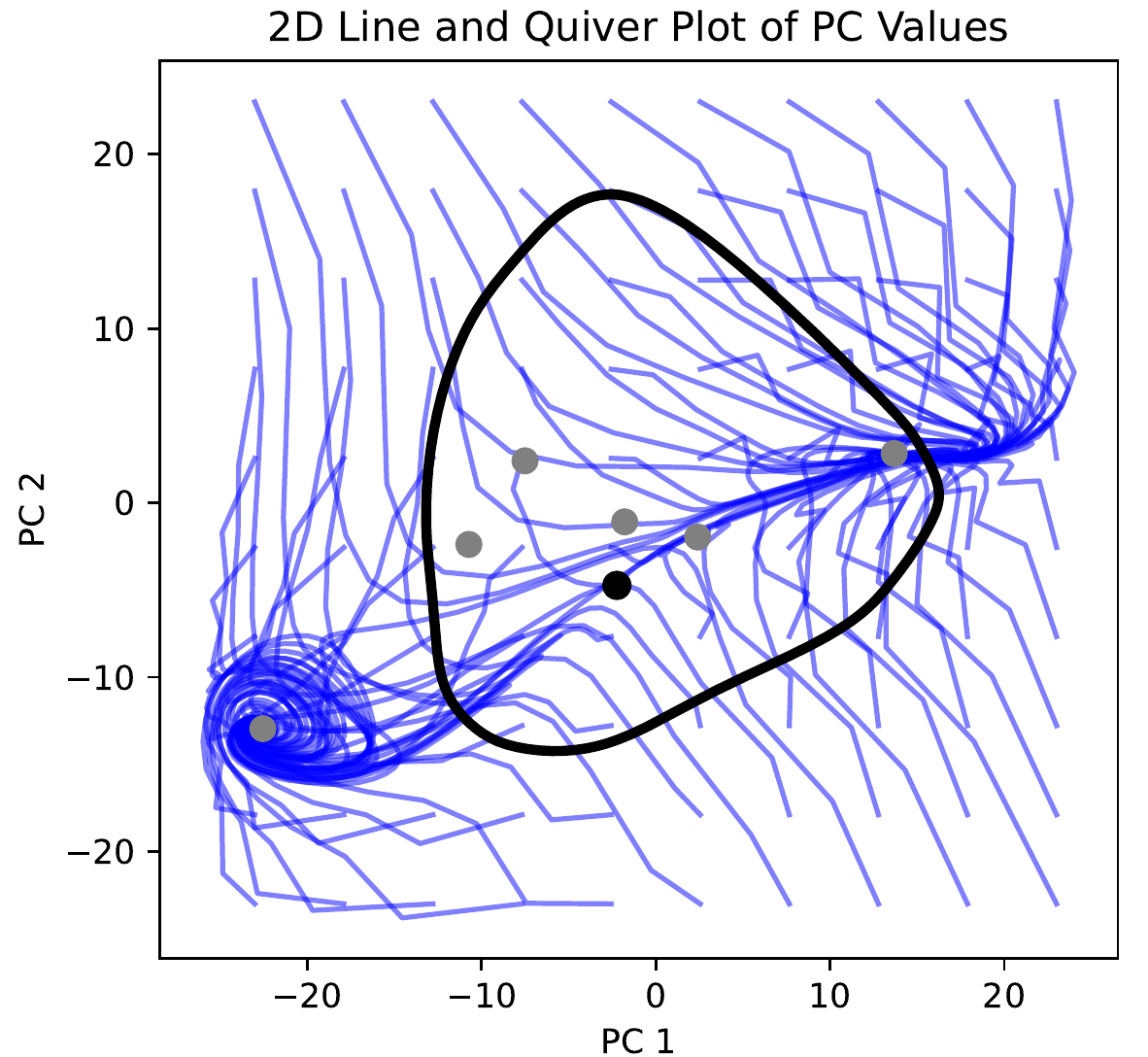} \\
    
    \includegraphics[valign=c,width=0.35\textwidth]{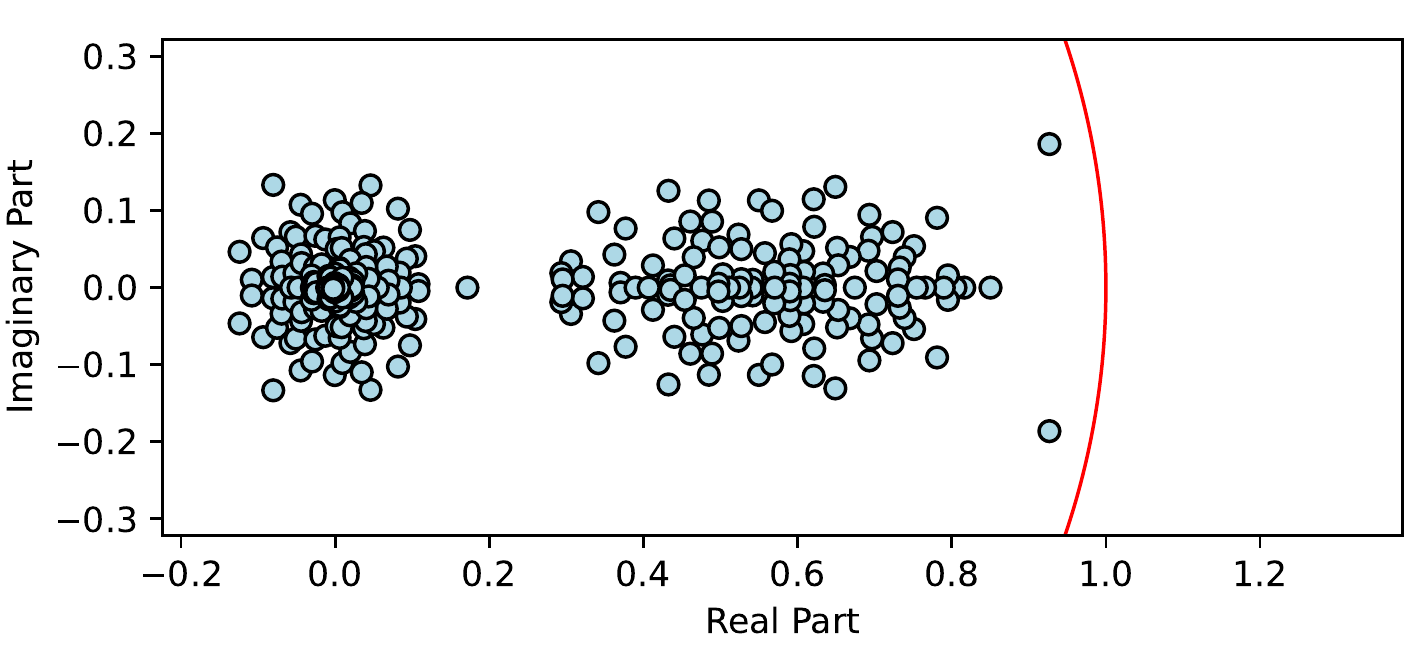}
    \includegraphics[valign=c,width=0.225\textwidth]{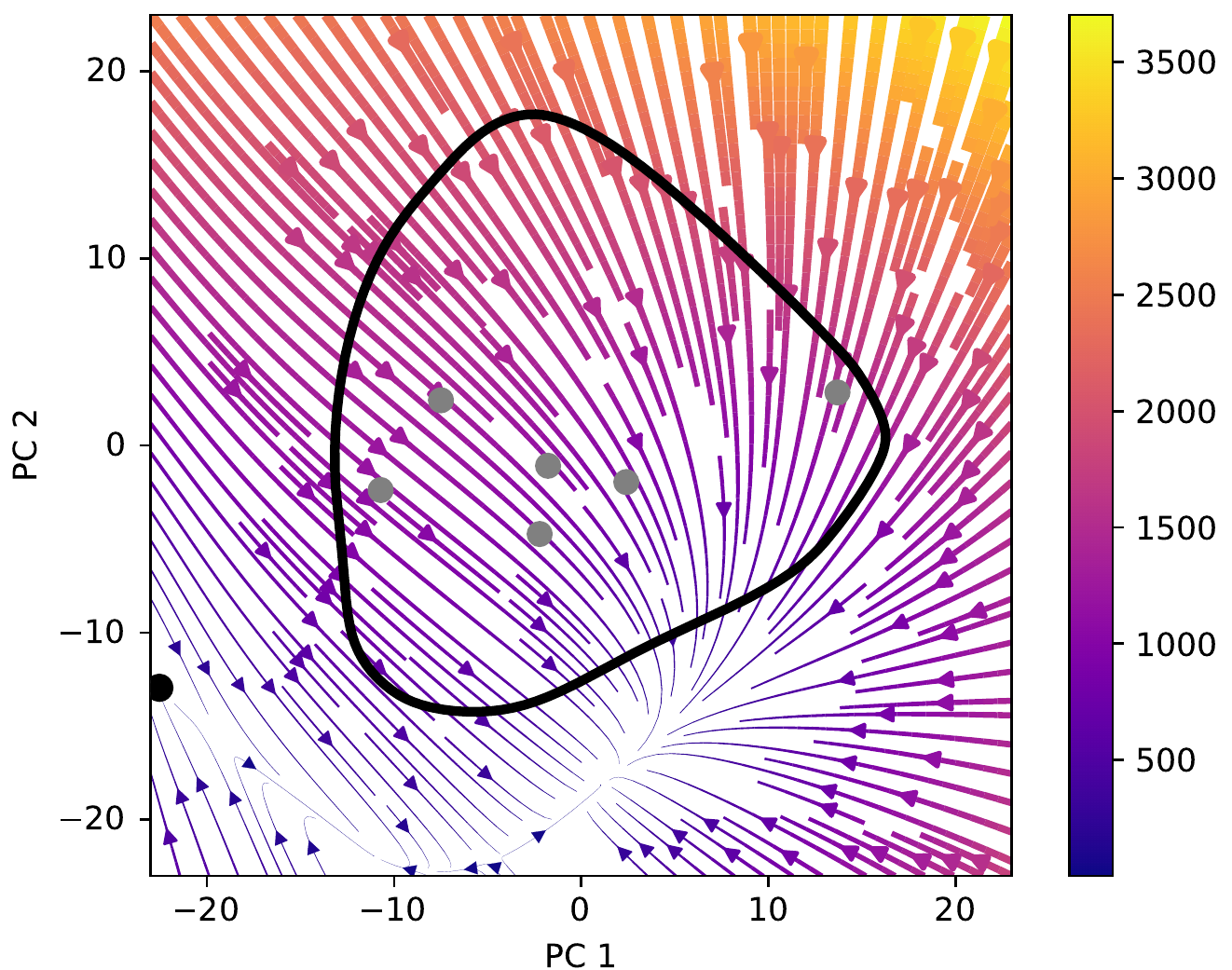}
    \includegraphics[valign=c,width=0.225\textwidth]{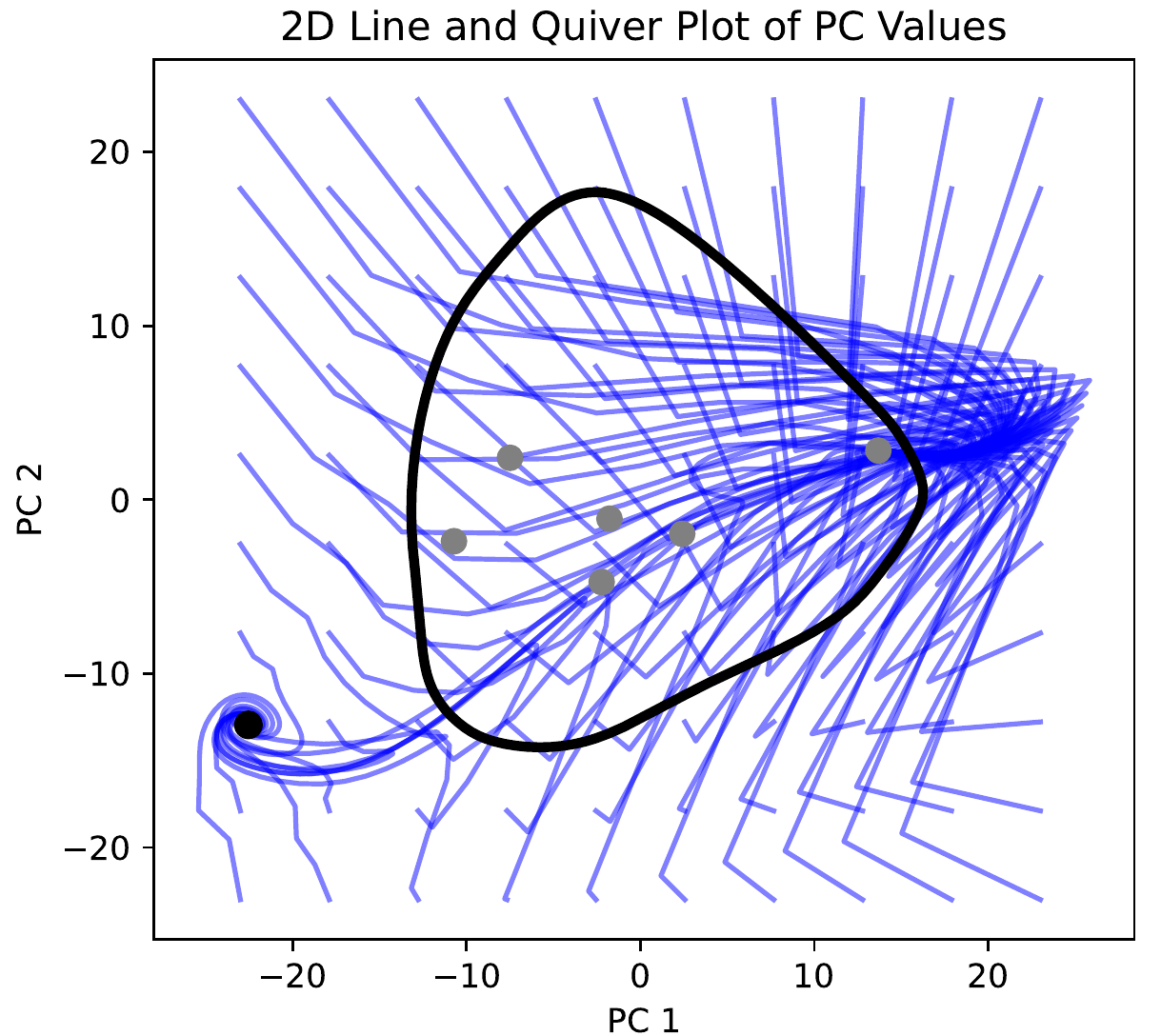} \\
    
    \includegraphics[valign=c,width=0.35\textwidth]{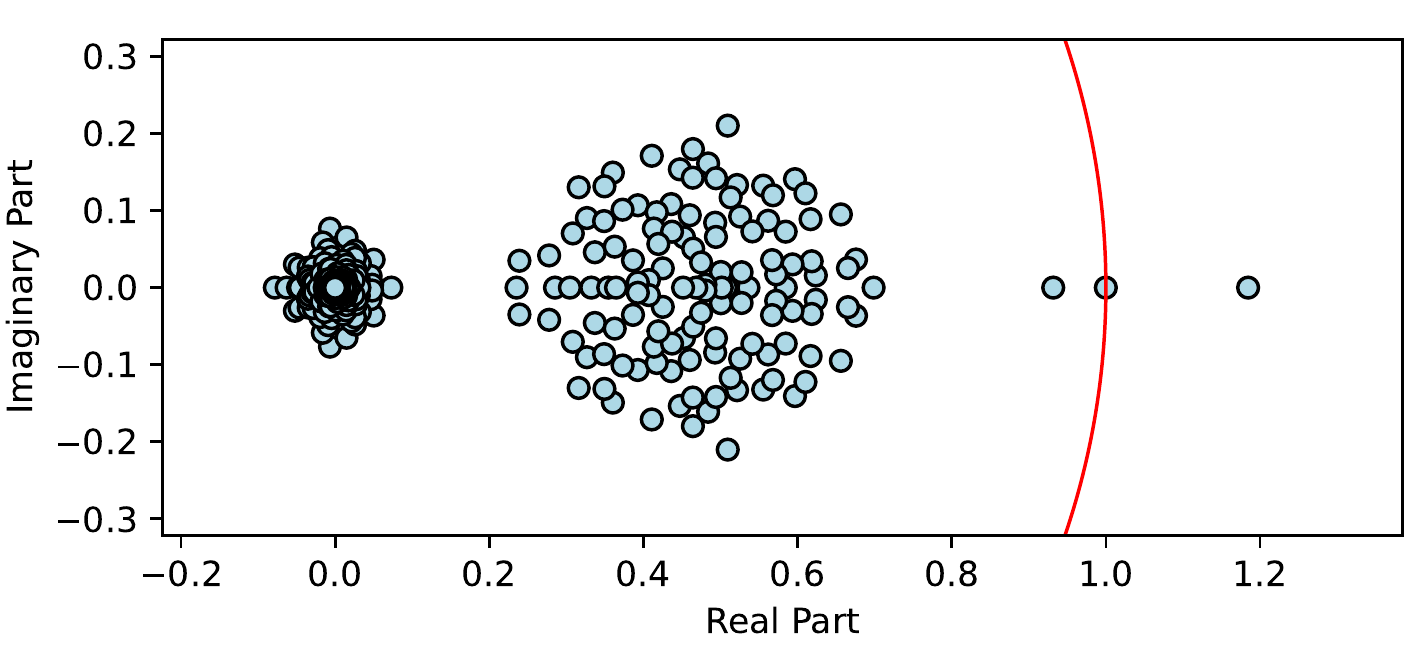}
    \includegraphics[valign=c,width=0.225\textwidth]{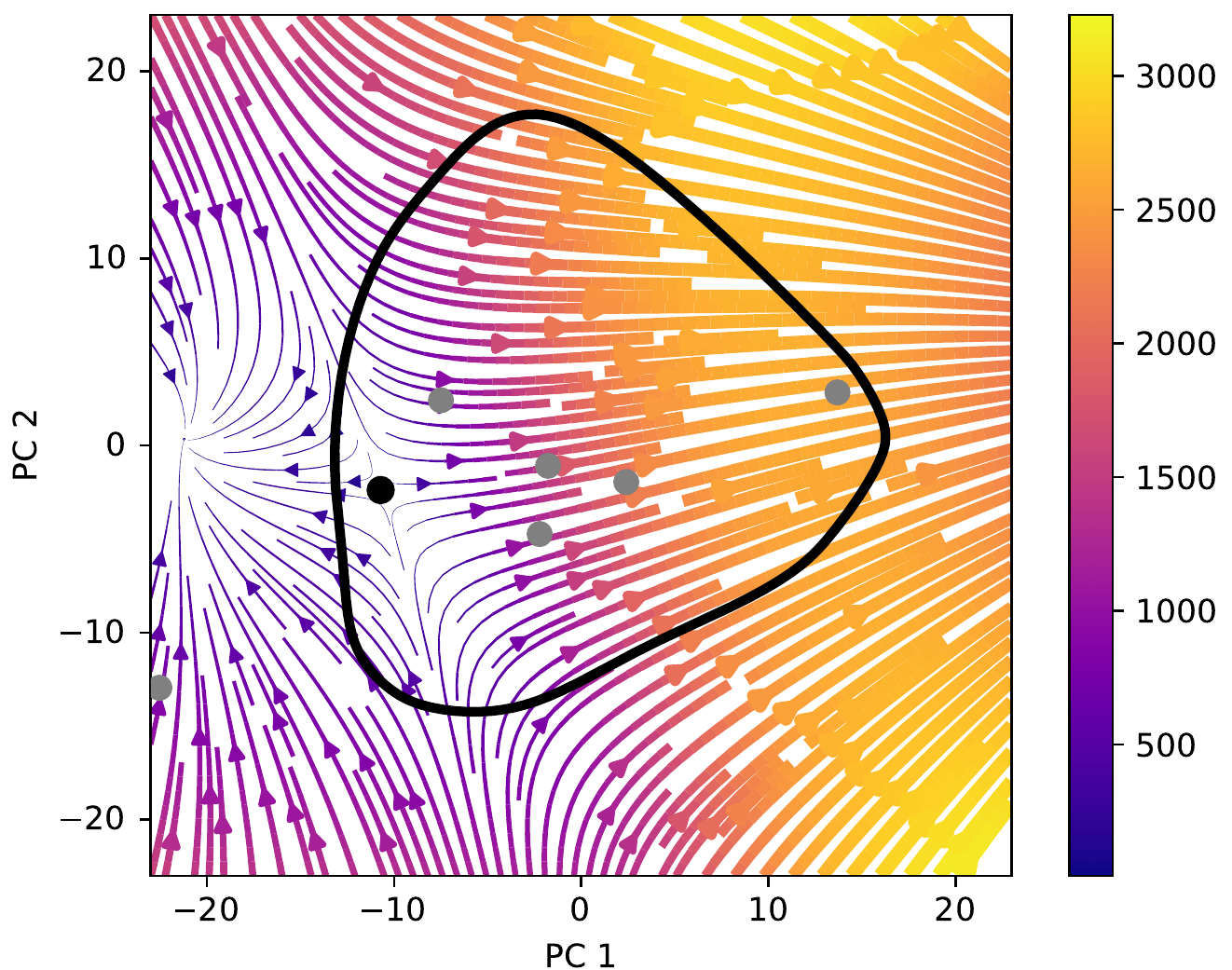}
    \includegraphics[valign=c,width=0.225\textwidth]{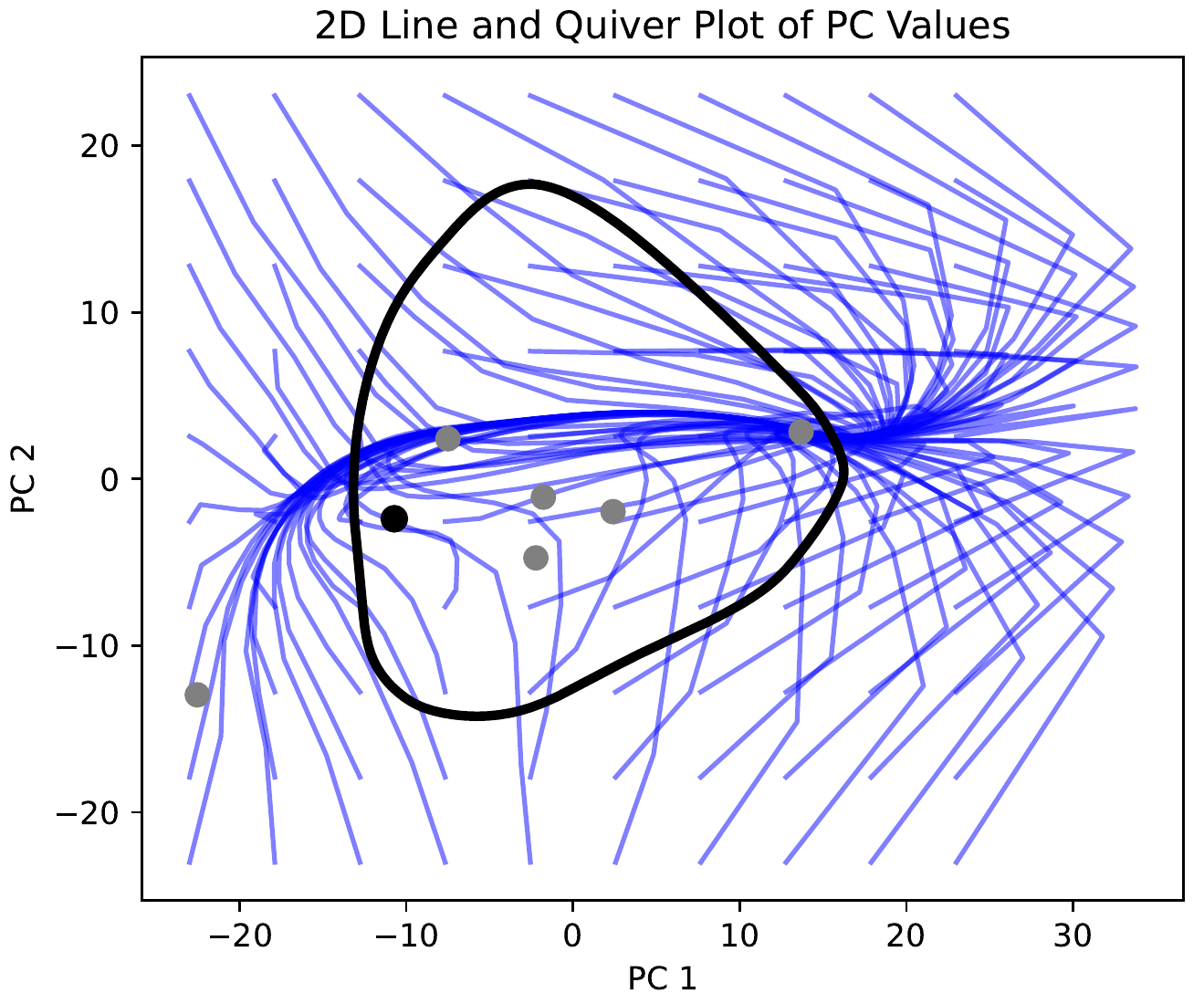} \\
    
    \caption{Companion to Figure \ref{fig:fixed_points_LSTM16}, cataloguing all seven fixed points of LSTM4 model.}
    \label{fig:supp_matl_fixed_points_LSTM4}
\end{figure}

All seven fixed points of LSTM4 models are catalogued in Figure \ref{fig:supp_matl_fixed_points_LSTM4}.

    
    
    

\newpage

\begin{figure}[h!]
    \centering
    \includegraphics[width=0.5\textwidth]{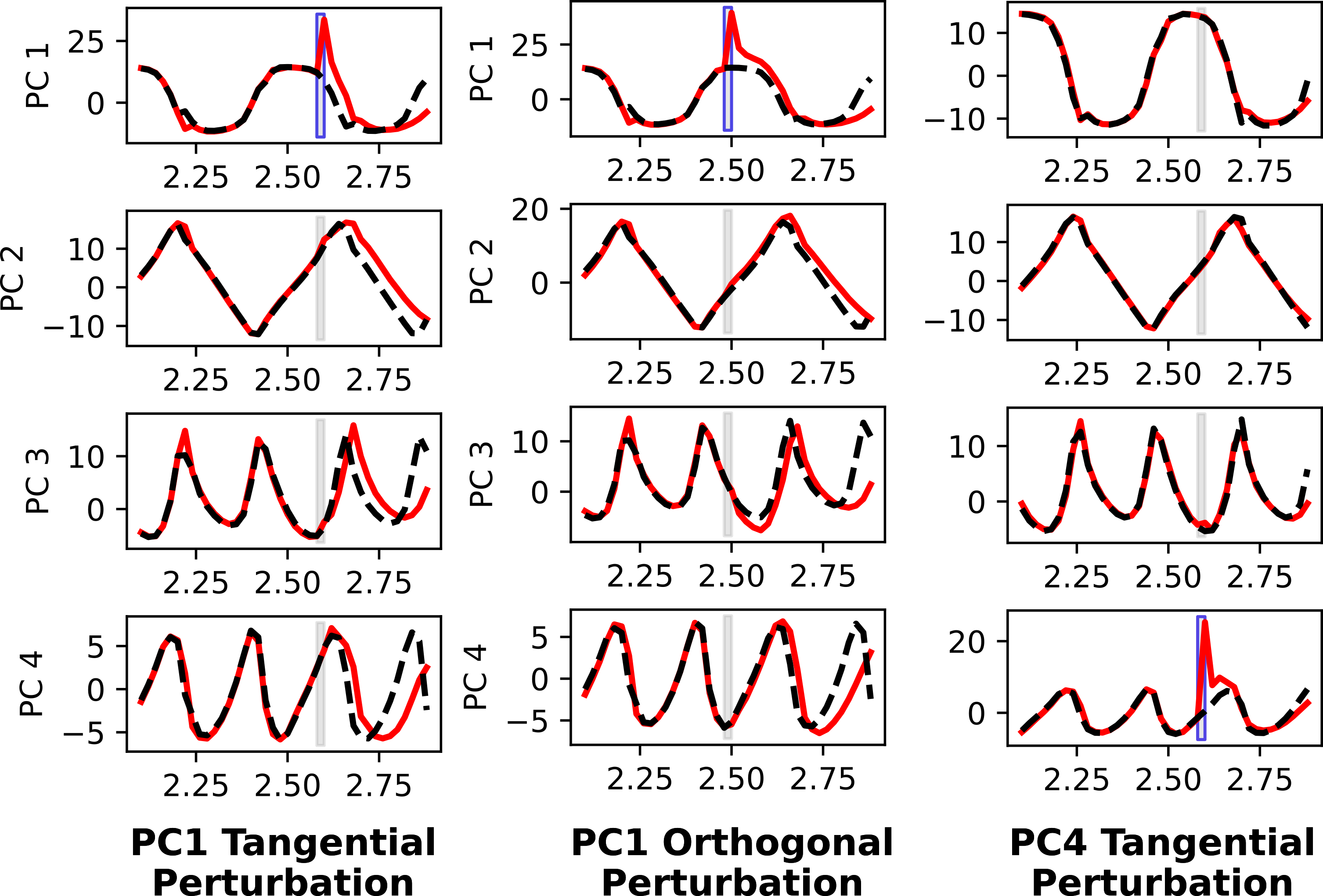}
    \caption{Companion to Figure \ref{fig:neural_hc_pc_perturb_LSTM4}, illustrating the response of recurrent states in PC1 through PC4 directions, before and after targeted neural perturbations.  all seven fixed points of LSTM4 model.}
    \label{fig:neural_hc_pc_perturb_LSTM16}
\end{figure}

The perturbation response of LSTM16 during forward walking is shown in Figure \ref{fig:neural_hc_pc_perturb_LSTM16}.

\end{document}